\documentclass[preprint,12pt]{elsarticle}

\usepackage{geometry}
\renewcommand{\paragraph}[1]{%
	\par 
	\vspace{1ex} 
	\noindent 
	\textit{#1} 
	\par 
	\vspace{0.5ex} 
}

\usepackage{amssymb}
\usepackage{amsmath}

\usepackage{algorithm}
\usepackage{algpseudocode}
\usepackage{tcolorbox}
\usepackage{booktabs}
\usepackage{subcaption}
\usepackage{xcolor}

\usepackage[numbers]{natbib}
\journal{Robotics and Computer-Integrated Manufacturing}

\begin{document}

\begin{frontmatter}
	
	\title{VLM-DEWM: Dynamic External World Model for Verifiable and Resilient Vision-Language Planning in Manufacturing} 
	
	\author[1]{Guoqin Tang}
	\author[1]{Qingxuan Jia} 
	\author[1]{Gang Chen\corref{cor1}}  
	\ead{buptcg@163.com}  
	\author[1]{Tong Li}
	\author[1]{Zeyuan Huang}
	\author[1]{Ning Ji}
	\author[1]{Zihang Lv}

	\address[1]{School of Intelligent Engineering and Automation, Beijing University of Posts and Telecommunications, Beijing, 100876, China}
	
	\cortext[cor1]{Corresponding author}
	
\begin{abstract}
Vision-language model (VLM) shows promise for high-level planning in smart manufacturing, yet their deployment in dynamic workcells faces two critical challenges: (1) stateless operation—they cannot persistently track out-of-view states, causing world-state drift; and (2) opaque reasoning—failures are difficult to diagnose, leading to costly blind retries. This paper presents VLM-DEWM, a cognitive architecture that decouples VLM reasoning from world-state management through a persistent, queryable Dynamic External World Model (DEWM). Each VLM decision is structured into an Externalizable Reasoning Trace (ERT)—comprising action proposal, world belief, and causal assumption—which is validated against DEWM before execution. When failures occur, discrepancy analysis between predicted and observed states enables targeted recovery instead of global replanning. We evaluate VLM-DEWM on multi-station assembly, large-scale facility exploration, and real-robot recovery under induced failures. Compared to baseline memory-augmented VLM systems, VLM-DEWM improves state-tracking accuracy from 56\% to 93\%, increases recovery success rate from below 5\% to 95\%, and significantly reduces computational overhead through structured memory. These results establish VLM-DEWM as a verifiable and resilient solution for long-horizon robotic operations in dynamic manufacturing environments.
\end{abstract}


\end{frontmatter}



\section{Introduction}
\label{sec:introduction}
Flexible manufacturing systems increasingly rely on autonomous robots to execute long-horizon tasks such as multi-station assembly, inter-station material handling, and inventory management~\cite{r1-smartmanu}. Unlike single-station operations, these tasks span multiple work areas where the workspace changes dynamically due to inventory replenishment, human intervention, occlusions, and execution disturbances. Reliable operation therefore requires long-horizon planning across stations, persistent tracking of out-of-view states, and fast failure diagnosis and recovery to minimize downtime.

Pre-trained vision--language models (VLMs) offer strong semantic understanding and task decomposition capabilities, making them attractive for high-level robotic planning. However, VLM-based planners face a fundamental responsibility mismatch in dynamic manufacturing workcells: they are required to both reason semantically \textit{and} maintain long-horizon world states within a transient context window. This coupling is inherently fragile—cross-station consistency degrades under occlusion and inventory changes, leading to world-state drift over time~\cite{r8-lack3d,r9-leak3d}. Moreover, when execution fails due to stale beliefs, the system lacks interpretable root-cause explanations~\cite{r11-lack3d}, and recovery degenerates into blind retries or expensive global replanning, increasing cycle time and downtime.

Consider a representative multi-station task: a mobile manipulator retrieves a Type-1 part from Station~A, transports it to Station~C, and completes assembly. The robot's memory indicates one Type-1 part at Station~A, but upon arrival perception detects two identical parts due to replenishment. Without resolving this memory--observation 
conflict, the robot may pick the wrong part or fail to terminate. If the part is later dropped at an unobserved location, recovery further requires diagnosing both the failure cause (e.g., grasp loss, collision, or actuator fault) and the updated part location; otherwise the system again resorts to blind retries or global replanning.

In isolation, each of these issues—memory staleness, observation conflict, undiagnosed failure—might be recoverable through local corrections. However, long-horizon tasks amplify their impact in ways that single-station operations do not. First, errors accumulate: a single unresolved memory--observation conflict propagates incorrect beliefs to all downstream decisions, causing cascading failures that compound across subsequent stations. Second, retry costs escalate non-linearly: each blind retry consumes cycle time without addressing the root cause, and over a 50-step task, even a 5\% per-step failure rate yields near-zero end-to-end success ($0.95^{50} \approx 8\%$). Third, diagnosis grows intractable: as the action history lengthens, identifying which step introduced the error becomes increasingly difficult without structured state tracking. These compounding effects point to three requirements for reliable long-horizon VLM planning: state persistence across stations and occlusions, decision verifiability to arrest error propagation, and failure diagnosability to enable efficient recovery.

Rather than relying on larger context windows, we address this mismatch through architectural separation: persistent state maintenance is delegated to an external, queryable world model, while the VLM focuses solely on semantic reasoning over structured state representations.This separation yields two design principles. First, decouple state from reasoning: maintain an explicit, queryable world state that is continuously synchronized with perception and remains accessible across station transitions, instead of implicitly encoding state in the context window. Second, make decisions verifiable and diagnosable: externalize VLM outputs into auditable artifacts, validate them against physical consistency before execution, and compare them with post-execution outcomes to enable informed recovery.

Based on these principles, we propose VLM-DEWM (VLM with Dynamic External World Model), a cognitive architecture operating on a ``Database--Transaction--Verification'' paradigm. DEWM maintains a persistent, queryable database of semantic relations, 3D geometry, 
and execution history.\footnote{Our ``world model'' refers to a persistent state database 	for maintaining the current world state, distinct from learned dynamics models in model-based RL that predict $P(s'|s,a)$, or manually engineered PDDL domains that encode symbolic preconditions. DEWM does not model dynamics; it maintains a synchronized snapshot that supports structured, verifiable zero-shot task decomposition.} On top of DEWM, an ``Externalizable Reasoning Trace (ERT)'' structures each VLM decision into an action proposal, a world belief, and a causal assumption, all of which are validated against physical consistency before execution. A closed-loop cognitive pipeline synthesizes task-specific queries from DEWM, audits VLM outputs through multi-layer checks, and performs targeted failure diagnosis when discrepancies arise.

We evaluate VLM-DEWM in simulation and real-robot experiments across three manufacturing-relevant scenarios: multi-station assembly under occlusions, large-scale facility exploration with spatial queries, and dynamic recovery under induced execution failures. Compared to state-of-the-art memory-augmented VLM systems, VLM-DEWM improves 
state-tracking accuracy from 56\% to 93\% in exploration tasks, achieves 95\% recovery success rate in settings where the baselines rarely succeed, and reduces VLM inference overhead by over 70\%. These results demonstrate that architectural separation of state maintenance from semantic reasoning substantially improves robustness for long-horizon robotic operations in dynamic manufacturing environments.

The main contributions are:
\begin{enumerate}
	\item \textbf{Dynamic External World Model (DEWM):} Resolves world-state drift by decoupling state maintenance from reasoning, ensuring persistent cross-station memory despite occlusions and transient context limitations.
	
	\item \textbf{Externalizable Reasoning Trace (ERT):} Mitigates opaque reasoning by structuring VLM decisions into verifiable artifacts, enabling pre-execution physical checks to arrest error propagation.
	
	\item \textbf{Discrepancy-Driven Diagnosis:} Eliminates costly blind retries via $CS$-anchored discrepancy analysis, which leverages prediction-observation mismatches to enable precise, targeted recovery.
\end{enumerate}

\section{Related Work}
\label{sec:related_work}

As established in Section~\ref{sec:introduction}, long-horizon robotic manipulation poses three core challenges: \textbf{state persistence} across 
occlusions and station transitions, \textbf{decision verifiability} for auditable execution, and \textbf{failure diagnosability} for efficient recovery. We review related work through these three lenses, examining how existing 
approaches in VLM-driven planning (Section~\ref{sec:rw_vlm_planning}), memory-augmented agents (Section~\ref{sec:rw_memory_augmented}), and scene representations (Section~\ref{sec:rw_scene_representation}) address—or fail to address—each challenge.

\subsection{VLM-Driven Robot Planning: From Generalization to Auditable Execution}
\label{sec:rw_vlm_planning}

Research in VLM-driven planning generally falls into two paradigms: joint training (e.g., PaLM-E~\cite{r12-Palm-E}, PaLi-X~\cite{r13-PaLi-x}) and frozen deployment (e.g., RT-2~\cite{r15-rt2}, OpenVLA~\cite{ref-openvla}). The latter has recently expanded to scalable transformers such as $\pi_0$~\cite{ref-pi0} and RDT-1B~\cite{ref-rdt}, enabling end-to-end visuomotor policies that operate closer to the control interface.

While these methods demonstrate impressive generalization, they face a critical bottleneck in manufacturing: state accountability. In production environments, execution must be auditable: when an assembly task fails (e.g., insertion jamming), operators need to trace whether the error arose from perceptual occlusion, geometric drift, or an invalid logical precondition. 
However, end-to-end models encode state implicitly in transient contexts, making root-cause attribution ill-posed without explicit state variables. Strategies like Inner Monologue~\cite{ref-inner} or ProgPrompt~\cite{ref-progprompt} attempt to externalize reasoning via textual logs, yet free-form language does not guarantee the metric determinism required for digital twins or MES integration. Consequently, purely VLM-based planners remain ``open-loop'' with respect to state auditability.

\subsection{Memory-Augmented VLMs: Semantic Persistence without Metric Verification}
\label{sec:rw_memory_augmented}

The implicit state encoding discussed above manifests as \textit{state forgetting}—the inability to persistently track object states across occlusions and station transitions. To address this, recent works couple VLMs with external knowledge bases, exemplified by SAGE~\cite{ref-sage} and GEAR~\cite{ref-gear}, which employ Retrieval-Augmented Generation (RAG) to fetch context from semantic 
maps or vector databases.

While RAG improves persistence~\cite{r20}, it introduces a dual mismatch in geometric grounding and update latency. First, a fundamental geometric gap exists: RAG relies on vector-space similarity (e.g., cosine distance), which cannot verify metric spatial relations required for physical manipulation. Retrieving a statement like ``Part~A is aligned'' is insufficient to certify 
feasibility without explicit pose verification~\cite{ref-RAG-LLM-ROBOT}. Even graph-based variants~\cite{ref-graph-rag-3dprint} abstract away metric details 
(e.g., object poses, spatial offsets) essential for collision-free planning. Second, a dynamic mismatch persists: the standard ``update--index--retrieve'' loop often lags behind fast-changing workcell states, causing retrieved context to reflect outdated information and leading to execution failures~\cite{ref-kg-review}.

Agentic frameworks with self-correction (e.g., Corrective RAG~\cite{ref-c-rag}, Agent-G~\cite{ref-agent-rag}) partially address staleness but still lack geometry-grounded verification, limiting their applicability to long-horizon manipulation where persistent state tracking and metric validation are essential.

\subsection{Scene Representations: Bridging Geometry and Assembly-Process-State}
\label{sec:rw_scene_representation}

Scene representations in robotics must balance geometric fidelity with semantic reasoning. Current approaches struggle to satisfy both demands simultaneously.

On the geometric end, representations have evolved from discrete grids (OctoMap~\cite{ref-octomap}) to continuous neural fields 
(CATNIPS~\cite{ref-catnips}, BundleSDF~\cite{ref-bundleSDF}) and Gaussian splatting (GraspSplats~\cite{ref-graspsplat}). While achieving high visual fidelity, these methods typically treat assembly constraints as opaque optimization costs and lack explicit encoding of interaction phases—such as the transition from free-space motion to constrained insertion—creating barriers for failure-mode identification.

On the semantic end, 3D scene graphs (e.g., DovSG~\cite{ref-DOV}, 
RoboEXP~\cite{ref-roboExp}) encode logical relations and action affordances, while methods like RP-SG~\cite{ref-RP-SG} infer spatial relations under occlusion via graph networks. However, these approaches often discard metric details (e.g., precise 6-DoF poses) essential for physical feasibility verification and failure diagnosis.

Existing frameworks rarely capture the complete picture: they either track geometry without process-aware semantics, or encode task logic without reliable metric grounding. VLM-DEWM addresses this gap through a hybrid $\mathcal{S}$-$\mathcal{G}$ architecture that couples high-fidelity geometric tracking with diagnosable, process-aware state abstraction—providing both the metric substrate for physical verification and the semantic interface for VLM reasoning.

\section{Methodology}
\label{sec:methodology}

As argued in Section~\ref{sec:introduction}, reliable VLM-based planning requires separating persistent state maintenance from semantic reasoning. We instantiate this principle through VLM-DEWM, a framework operating on a ``Database--Transaction--Verification paradigm'': the DEWM serves as a persistent database maintaining world state, the VLM outputs an Externalizable Reasoning Trace (ERT) as a verifiable transaction, and a Verification Engine audits the ERT against DEWM before execution.

Formally, the DEWM state at time $t$ is defined as:
\begin{equation}
	\text{DEWM}_t = \langle \Omega_t, M_t, CS_t \rangle
\end{equation}
where $\Omega_t = \langle \mathcal{S}, \mathcal{G}, \mathcal{L} \rangle$ is the environment memory core (spatial network, semantic graph, and shape prior library), $M_t$ is the task memory tracking execution progress, and $CS_t$ is the constraint state anchoring the robot's interaction phase for failure diagnosis.

\begin{figure*}[t]
	\centering
	\includegraphics[width=0.85\linewidth]{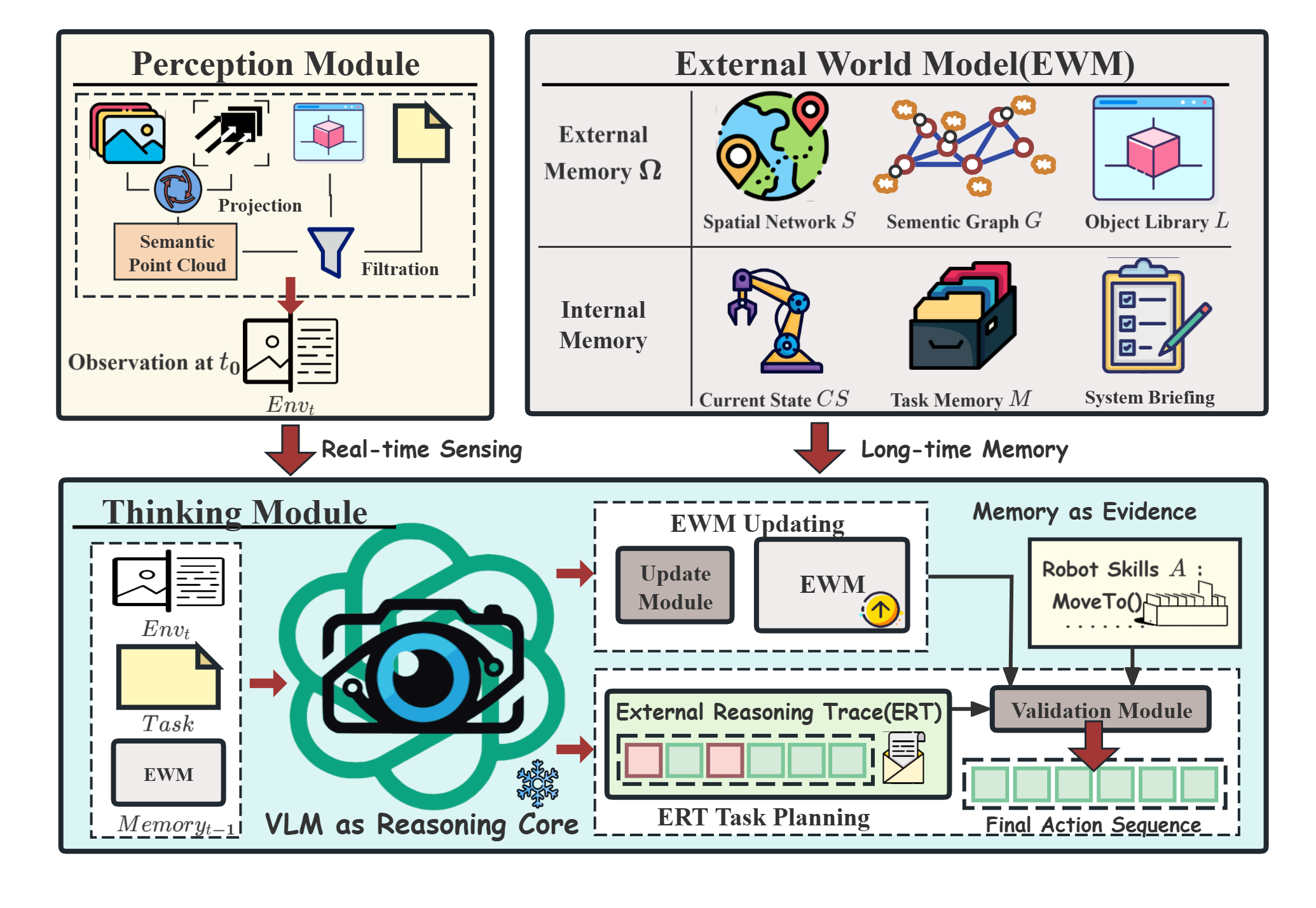} 
	\caption{\textbf{Architectural Overview of the VLM-DEWM Framework.} 
		Operating on a Database-Transaction paradigm, the system is orchestrated through three tightly coupled modules. 
		The perception module first projects raw multimodal sensory streams into a structured instantaneous snapshot ($Env_t$) via geometric projection and filtration. 
		This snapshot drives the thinking module, where the VLM serves as the reasoning core to simultaneously update the external world model (DEWM)—the central knowledge hub comprising environmental ($\Omega$) and internal ($M, CS$) memory—and synthesize a structured \textit{External Reasoning Trace (ERT)}. 
		Crucially, the validation module treats the persistent memory in the DEWM as ground-truth evidence to rigorously verify the ERT's logical and physical consistency before dispatching the final action sequence to the Robot Skill Library $\mathcal{A}$ for execution.}
	\label{fig:system_overview}
\end{figure*}

As illustrated in Fig.~\ref{fig:system_overview}, this formalism is instantiated as a closed-loop system with four modules: \textit{Perception}, \textit{DEWM State Management}, \textit{VLM-based Reasoning}, and \textit{Validation}. 
At each step, the system first perceives raw observations $O_t$ and updates the world state $\Omega$ together with an interaction-stage state $CS$. 
It then queries $\Omega$ to construct task-relevant context for the VLM, which generates an ERT hypothesis for the next subtask. 
Before dispatching actions to the primitive skill library $\mathcal{A}$, the validation module checks the ERT against $\Omega$ and blocks geometrically infeasible plans; 
after execution, the system compares the observed outcome with the ERT's causal assumption to trigger diagnosis and informed replanning when discrepancies occur.

\subsection{The Environment Memory Core ($\Omega$): A Hybrid Semantic-Geometric Representation}
\label{sec:environment_memory}

\begin{figure*}[t]
	\centering
	\includegraphics[width=\linewidth]{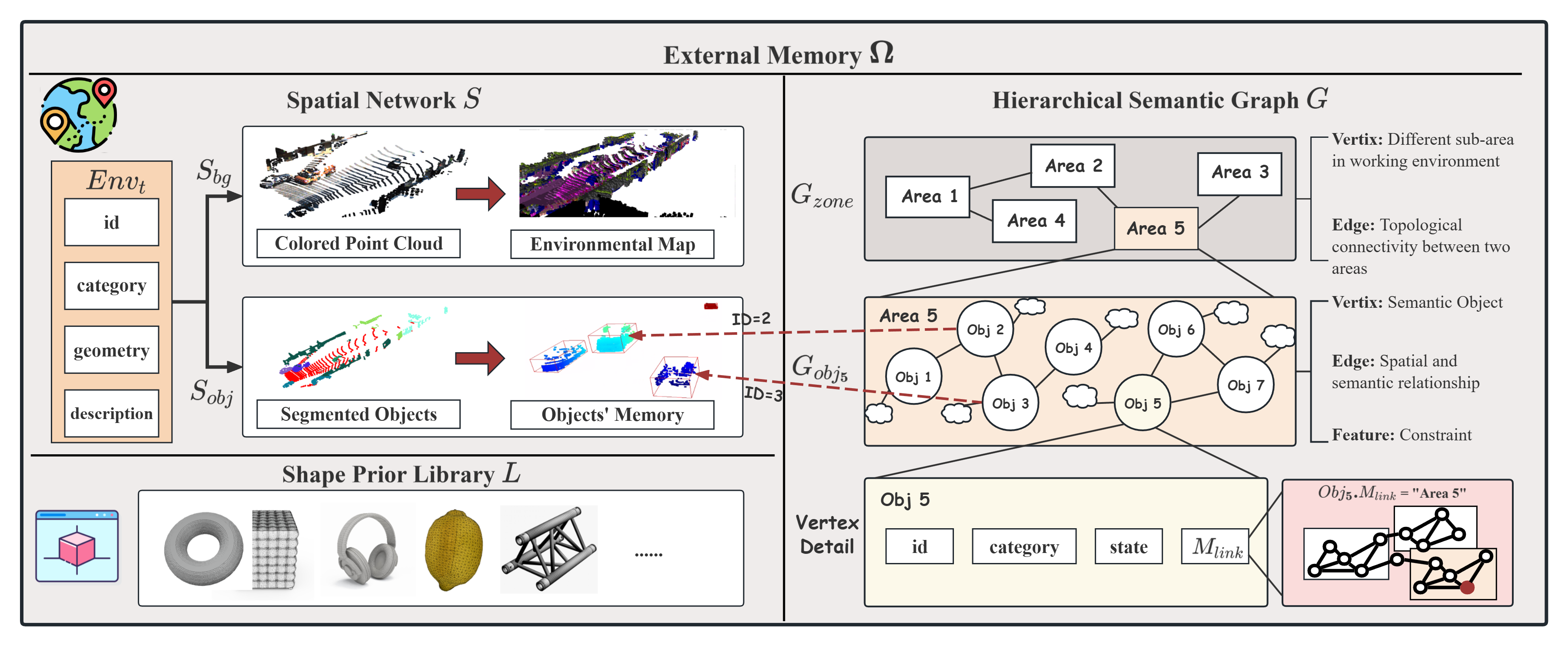} 
	\caption{\textbf{Structural Composition of the Environment Memory Core ($\Omega$).} 
		The Spatial Network ($\mathcal{S}$) maintains geometric state (left), while the Hierarchical Semantic Graph ($\mathcal{G}$) encodes logical relations (right). The Grounding Link ($\Phi_{ground}$, red) anchors every semantic node to its physical instantiation.}
	\label{fig:dewm_structure}
\end{figure*}

The Dynamic External World Model (DEWM) serves as the system's knowledge foundation. Its primary component, the Environment Memory Core $\Omega$, is designed based on the philosophy of \textit{structural decoupling}: separating high-level semantic logic from low-level geometric precision.
Formally, we define $\Omega$ as a tripartite tuple:
\begin{equation}
	\Omega = \langle \mathcal{S}, \mathcal{G}, \mathcal{L} \rangle
\end{equation}
where $\mathcal{S}$ represents the physical state, $\mathcal{G}$ encapsulates logical relations, and $\mathcal{L}$ provides shape priors (see Fig.~\ref{fig:dewm_structure}).

In this architecture, $\mathcal{G}$ provides a queryable interface for VLM reasoning, $\mathcal{S}$ serves as the metric substrate for feasibility checking and motion planning, and $\mathcal{L}$ supplies on-demand high-fidelity geometry for robust 6-DoF registration and occlusion-robust shape inference. All downstream modules interact with $\Omega$ via a \textit{semantic-to-metric grounding path}: semantic queries on $\mathcal{G}$ are resolved to geometric entities in $\mathcal{S}$ via explicit links, enabling verification against metric constraints.

\subsubsection{The Geometric Layer: Spatial Network ($\mathcal{S}$)}
The Spatial Network $\mathcal{S}$ maintains the probabilistic state estimate of the physical state of the environment. To balance global planning efficiency with local manipulation precision, we adopt a dual-representation strategy: $\mathcal{S} = \langle \mathcal{S}_{bg}, \mathcal{S}_{obj} \rangle$.

This design emulates human spatial memory—\textit{``focus on foreground, blur the background''}—where task-relevant objects receive high-fidelity tracking while static structures are compressed into occupancy grids. Specifically, entities segmented with task-relevant labels (Section~\ref{sec:data_association}) are registered in $\mathcal{S}_{obj}$, while unassociated geometric primitives are fused into $\mathcal{S}_{bg}$.

\begin{enumerate}
	\item Global Background Map ($\mathcal{S}_{bg}$):
	We utilize a coarse-grained, semantically labeled probabilistic voxel grid to represent static structures (e.g., walls, workbenches). This provides the free-space information necessary for global collision avoidance and coarse motion planning.
	
	\item Dynamic Object Map ($\mathcal{S}_{obj}$): 
	This is an instance-indexed registry tracking all interactive entities. For each tracked object $i$, its geometric state $s_{obj}^i$ is defined as:
	\begin{equation}
		s_{obj}^i = \langle \text{id}_i, \mathbf{T}_i, \mathcal{E}_i \rangle
	\end{equation}
	Here, $\text{id}_i$ is a pointer to the Shape Library $\mathcal{L}$; $\mathbf{T}_i \in SE(3)$ denotes the precise 6-DoF pose updated via multi-frame tracking; and $\mathcal{E}_i(\boldsymbol{\mu}_i, \boldsymbol{\Sigma}_i)$ represents a Gaussian Envelope.
\end{enumerate}

\textit{Remark:} The Gaussian envelope $\mathcal{E}_i$ acts as a \textit{probabilistic spatial index} for rapid VLM queries (e.g., ``near X'') and candidate pruning, rather than for fine-grained contact modeling. When high-precision pose estimation is required (e.g., for grasping or insertion), precise geometry is retrieved on-demand from $\mathcal{L}$ via the shape prior link.
	
\subsubsection{The Semantic Layer: Hierarchical Semantic Graph ($\mathcal{G}$)}
The semantic layer $\mathcal{G}$ serves as the interface for VLM reasoning. We construct a Hierarchical Semantic Graph (HSG) to support multi-resolution planning: $\mathcal{G} = \langle \mathcal{G}_{zone}, \mathcal{G}_{obj} \rangle$.

\begin{itemize}
	\item Zone Graph ($\mathcal{G}_{zone}$): Acts as the spatial root, where nodes represent topological regions (e.g., \texttt{Workbench\_A}) and edges denote reachability, allowing the planner to efficiently prune the search space.
	
	\item Object Graph ($\mathcal{G}_{obj}$): Modeled as a graph $\mathcal{G}_{obj}=(\mathcal{V}, \mathcal{E})$, where each vertex $v_i \in \mathcal{V}$ represents an object instance:
	\begin{equation}
		v_i = \langle \text{uid}_i, \text{label}_i, \text{state}_i, \text{Attr}_i, \Phi_{ground} \rangle
	\end{equation}
	Here, $\text{Attr}_i$ stores feature attributes extracted from VLM perception (e.g., ``fragile'', ``stack\_on \_X''). The function $\Phi_{ground}: \mathcal{V} \to \mathcal{S}_{obj}$ explicitly links the semantic node to its geometric counterpart in $\mathcal{S}$, ensuring that every abstract concept is physically grounded. 
	Edges $e_{ij} \in \mathcal{E}$ encode spatial relations (e.g., \texttt{On}) or affordances, derived from VLM inference. Note that these relations are stored as soft hypotheses and can be revised by the verification engine if inconsistent with geometric observations.
\end{itemize}

\subsubsection{The Shape Prior Library ($\mathcal{L}$): Digital-Twin-Aligned Priors}
The library $\mathcal{L}$ stores high-fidelity geometric priors derived from CAD models, providing digital-twin-aligned references for robust pose estimation and shape completion. In manufacturing settings, we pre-process CAD meshes into canonical point clouds associated with functional frames. These priors support robust 6-DoF registration under sparse visual features and occlusion-robust shape inference when objects are partially covered by the gripper. Detailed registration algorithms and occlusion handling strategies are provided in \ref{app:B1}.

\subsubsection{Dynamic Maintenance}
The mechanisms for updating $\Omega$ from perception—including data 
association, state fusion, and lifecycle management—are detailed in 
Section~\ref{sec:perception_dynamics}.

\subsubsection{Formal State Transition Semantics}
\label{sec:formal_semantics}
To ensure the deterministic maintenance of the world model, we define the DEWM update as a state transition function $\delta: (\Omega_t, CS_t) \times \pi_{act} \to (\Omega_{t+1}, CS_{t+1})$. 
This transition is designed to approximate database-style atomicity: 
if the rollback trigger is activated, the state reverts to $(\Omega_t, CS_t)$.
Specifically, for a primitive action $\pi_{act}$, the update is an atomic transaction defined in Table~\ref{tab:transition_rules}.

\begin{table}[h]
	\centering
	\caption{Atomic State Transition Rules for Primitive Skills}
	\label{tab:transition_rules}
	\small 
	\setlength{\tabcolsep}{3pt} 
	\begin{tabular}{l p{3.5cm} p{5.5cm} p{4cm}}
		\toprule
		\textbf{Action} $\pi$ & \textbf{Pre-Condition} ($\mathcal{G}$) & \textbf{Atomic Effects} ($\Omega_{t+1}, CS_{t+1}$) & \textbf{Rollback Trigger} \\
		\midrule
		\texttt{Pick(obj)} & $\exists \texttt{Clear(obj)}$ & 
		1. $\mathcal{S}_{obj}$: Detach $\mathcal{PC}_{obj}$ \newline
		2. $\mathcal{G}$: Del $\texttt{On(obj, *)}$ \newline
		3. $CS$: Set $\texttt{Holding(obj)}$ & Gripper width $>$ limit (Slip) \\
		\midrule
		\texttt{Place(loc)} & $CS=\texttt{Holding(obj)}$ & 
		1. $\mathcal{S}_{obj}$: Anchor at $loc$ \newline
		2. $\mathcal{G}$: Add $\texttt{On(obj, loc)}$ \newline
		3. $CS$: Set $\texttt{Idle}$ & Force feedback $>$ limit (Collision) \\
		\midrule
		\texttt{Insert(A, B)} & $\exists \texttt{Aligned(A, B)}$ & 
		1. $\mathcal{G}$: Add $\texttt{Inserted(A, B)}$ \newline
		2. $CS$: Set $\texttt{Idle}$ & Torque $>$ limit / Stall \\
		\bottomrule
	\end{tabular}
	\vspace{2pt}
	\newline
	\footnotesize{\textit{Note:} If the Rollback Trigger is activated via low-level controller feedback (see \ref{app:force_feedback} for the specific force-torque thresholds and event mapping logic), the state reverts to $(\Omega_t, CS_t)$. This table presents representative primitives. The full skill library $\mathcal{A}$ includes additional primitives such as \texttt{Move(loc)}, \texttt{Rotate(obj, angle)}, and \texttt{Open/Close\_Gripper}, each with corresponding transition rules.}
\end{table}

\subsection{Perception and Memory Dynamics: Maintaining Model Veracity}
\label{sec:perception_dynamics}

\begin{figure*}[h]
	\centering
	\includegraphics[width=\linewidth]{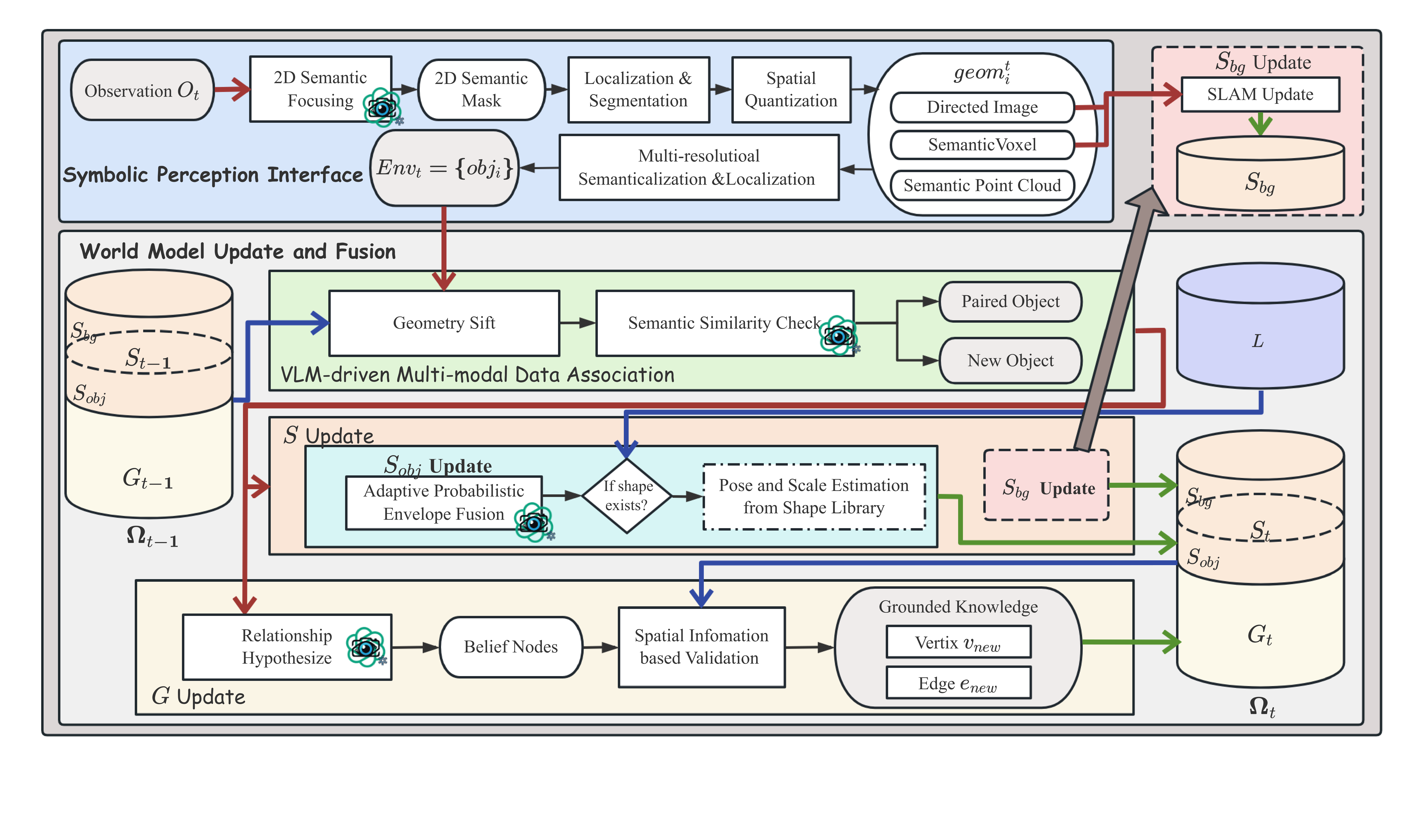} 
	\caption{\textbf{The Perception-to-Memory Synchronization Pipeline.} 
		Raw observations ($O_t$) are transformed into persistent belief updates via three stages: (1) task-focused perception producing $Env_t$, (2) geometric-semantic data association, and (3) dual-layer updates (continuous S-Update and discrete G-Update with geometric verification).}
	\label{fig:update_process}
\end{figure*}

While Section~\ref{sec:environment_memory} defined the static structure of the DEWM, this section details the dynamic processes that synchronize the internal model $\Omega$ with the evolving physical world. We employ a closed-loop pipeline ``\textit{Perceive $\rightarrow$ Associate $\rightarrow$ Fuse $\rightarrow$ Verify}'' to transform raw sensory streams into structured, physically grounded beliefs(see Fig.~\ref{fig:update_process}).

\subsubsection{Task-Focused Perception Interface}
Unlike passive perception, our system employs a VLM-driven attention mechanism. Given a task instruction, the VLM generates a set of \textit{Target Entities} and \textit{Regions of Interest}. Raw RGB-D streams $O_t$ are processed to extract geometric primitives only for these relevant targets, lifting 2D masks into 3D space. Formally, this produces an instantaneous snapshot $Env_t = \{ o_k \}_{k=1}^K$, where each observation $o_k$ contains a semantic label, a confidence score, and a distance-adaptive geometric abstraction—dense point clouds for near-field manipulation, voxels for mid-range collision checking, or billboard textures for far-field situational awareness (detailed in \ref{app:perception_pipeline}).

\subsubsection{VLM-Driven Data Association}
\label{sec:data_association}
Integrating new observations requires robust data association. Purely geometric matching fails in cluttered scenes, while purely semantic matching is computationally prohibitive. We therefore adopt a ``Filter-then-Adjudicate'' protocol that combines geometric pre-screening with VLM-assisted global assignment.

\paragraph{a. Geometric Gating (Filter):}

We prune infeasible pairs by computing the Mahalanobis distance $d_M$ between the observation's Gaussian envelope $(\boldsymbol{\mu}_{obs}, \boldsymbol{\Sigma}_{obs})$ and the memory entry's envelope $\mathcal{E}_i = (\boldsymbol{\mu}_{i}, \boldsymbol{\Sigma}_{i})$ defined in Section~\ref{sec:environment_memory}
\begin{equation}
	d_M(obs, i) = \sqrt{(\boldsymbol{\mu}_{obs} - \boldsymbol{\mu}_i)^\top (\boldsymbol{\Sigma}_{obs} + \boldsymbol{\Sigma}_i)^{-1} (\boldsymbol{\mu}_{obs} - \boldsymbol{\mu}_i)} .
\end{equation}
Pairs with $d_M \ge \tau_{geo}$ are rejected, avoiding physically impossible matches before semantic evaluation.

\paragraph{b. Semantic--Geometric Adjudication (Judge)}

On the remaining feasible pairs, we solve a global one-to-one assignment via the Hungarian algorithm using a hybrid cost:
\begin{equation}
	C_{ki} = \lambda_{iou} \bigl(1 - \text{IoU}(o_k, s_i)\bigr) + \lambda_{sem} \bigl(1 - \text{sim}(o_k, s_i)\bigr) .
\end{equation}
We implement geometric gating by masking infeasible edges, i.e., setting $C_{ki}=+\infty$ for pairs failing $d_M < \tau_{geo}$. Here, IoU is computed over 3D oriented bounding volumes (with a 2D fallback when depth is unreliable), and $\text{sim}(\cdot)$ denotes VLM-based semantic similarity. Matches must pass a final quality gate $C_{ki} \le \tau_{match}$. Unmatched observations enter a short confirmation window before being promoted to active entries, suppressing one-off false detections and preventing ghost artifacts.

All prompts, parsing rules, and association hyperparameters are provided in \ref{app:B1}.

\subsubsection{VLM-Modulated Gaussian Envelope Fusion} (S-Update)
\label{sec:adaptive_fusion}
Once associated, the state update is modulated by VLM reasoning. Standard filters degrade when semantic context affects reliability (e.g., transparent objects or gripper-induced occlusion). We therefore perform a semantically-adaptive Gaussian fusion.

For matched pairs, we maintain a Gaussian belief $(\boldsymbol{\mu}_t, \boldsymbol{\Sigma}_t)$ representing the object's spatial envelope. Given a new observation with parameters $(\boldsymbol{\mu}_{obs}, \boldsymbol{\Sigma}_{obs})$, we perform a weighted fusion in the information filter form:
\begin{align}
	\boldsymbol{\Sigma}_t^{-1} &= \boldsymbol{\Sigma}_{t-1}^{-1} + \gamma(\alpha_t) \cdot \boldsymbol{\Sigma}_{obs}^{-1} \label{eq:sigma_fusion}\\
	\boldsymbol{\mu}_t &= \boldsymbol{\Sigma}_t \left( \boldsymbol{\Sigma}_{t-1}^{-1} \boldsymbol{\mu}_{t-1} + \gamma(\alpha_t) \cdot \boldsymbol{\Sigma}_{obs}^{-1} \boldsymbol{\mu}_{obs} \right) \label{eq:mu_fusion}
\end{align}
where $\alpha_t \in \{\text{High, Medium, Low, Bad}\}$ is a discrete reliability judgment produced by the VLM from scene context (e.g., occlusion, lighting, and sensor quality), and $\gamma(\alpha_t)$ maps this judgment to a continuous weight factor that modulates the contribution of the new observation. 

This formulation treats the VLM as a meta-cognitive controller: when the VLM judges the observation as ``High'' reliability, $\gamma$ is large, allowing the new measurement to strongly influence the belief; when judged as ``Bad'' (e.g., due to heavy occlusion), $\gamma$ approaches zero, effectively ``freezing'' the belief at its prior value. This enables context-aware state estimation that traditional adaptive filters cannot achieve---for instance, adopting a conservative update strategy for fragile objects even when observation quality appears high.

The noise models, prompt templates, decoding/parsing rules, and the full mapping table used in all experiments are provided in \ref{app:B2}.

\subsubsection{Geometric Refinement and Background Maintenance}
For key objects requiring high-precision assembly, we perform a scale-aware robust ICP registration to align the current observation with the canonical Shape Prior $\mathcal{L}$, refining the pose estimate $\mathbf{T}_i \in SE(3)$ (details in \ref{app:B3}). Simultaneously, unassociated points are fused into the global background map $\mathcal{S}_{bg}$ using probabilistic occupancy updates, and open boundaries are handled via view-dependent billboard textures to provide visual context for navigation (details in \ref{app:B4}).

\subsubsection{Hypothesis-Verification Loop (G-Update)}
To maintain semantic consistency, relations proposed by the VLM (e.g., $On(A,B)$) are treated as hypotheses and verified against $\mathcal{S}$ at \textit{action boundaries}. For proximity-type relations (e.g., \texttt{Contact($A,B$)}), we validate consistency via a Chi-square test on Gaussian position estimates:
\begin{equation}
	(\boldsymbol{\mu}_B - \boldsymbol{\mu}_A)^\top (\boldsymbol{\Sigma}_A + \boldsymbol{\Sigma}_B)^{-1} (\boldsymbol{\mu}_B - \boldsymbol{\mu}_A) \le \chi^2_{thresh} .
\end{equation}
For support-type relations (e.g., \texttt{On($A,B$)}), we additionally apply conservative bounding-volume support checks with tolerance $\delta_{supp}$ as a proxy for physical support. The statistical basis, threshold selection, and verification criteria (including $\delta_{supp}$) for common relations are provided in \ref{app:B5}. This verification acts as a geometry-grounded firewall, preventing hallucinated relations from propagating into the planner's worldview.

\subsubsection{Long-Term Memory Curation}
\label{sec:memory_curation}
To prevent memory saturation in long-horizon tasks, we implement region-based confidence decay and lifecycle management. When the robot scans a zone $Z_k$ and fails to re-observe an expected entity $v_i$, its confidence score decays:
\begin{equation}
	c_t(v_i) = c_{t-1}(v_i) \cdot \lambda_{decay} .
\end{equation}
Entities transition through \texttt{Active}/\texttt{Uncertain}/\texttt{Archived} states based on confidence thresholds, and archived entities can be restored when re-observed. The complete lifecycle policy and curation algorithm are provided in \ref{app:B6}.

\subsubsection{Commit Conditions and Failure Handling}
The state transition is committed if and only if (1) the physical action is successfully executed by the controller (verified via force-torque feedback), and (2) the post-action perception verifies the expected geometric change specified in the ERT's \texttt{causal\_assumption} field (Section~\ref{sec:ert}). Specifically, for \texttt{Pick(obj)}, we verify that $obj$'s position now tracks the gripper frame; for \texttt{Place(loc)}, we verify $obj$ is stationary at $loc$ within tolerance $\epsilon_{place}$; for \texttt{Insert(A,B)}, we verify the relative pose $\mathbf{T}_{AB}$ matches the expected mating geometry.

If either condition fails (e.g., the object slips during transport), 
the transaction is rolled back and the Diagnosis Module 
(Section~\ref{sec:deep_diagnosis}) is triggered, preventing the 
``hallucination accumulation'' common in open-loop VLM planners.

\subsection{VLM-Guided Planning, Execution, and Recovery}
\label{sec:planning_execution}

\begin{figure*}[t]
	\centering
	\includegraphics[width=0.9\linewidth]{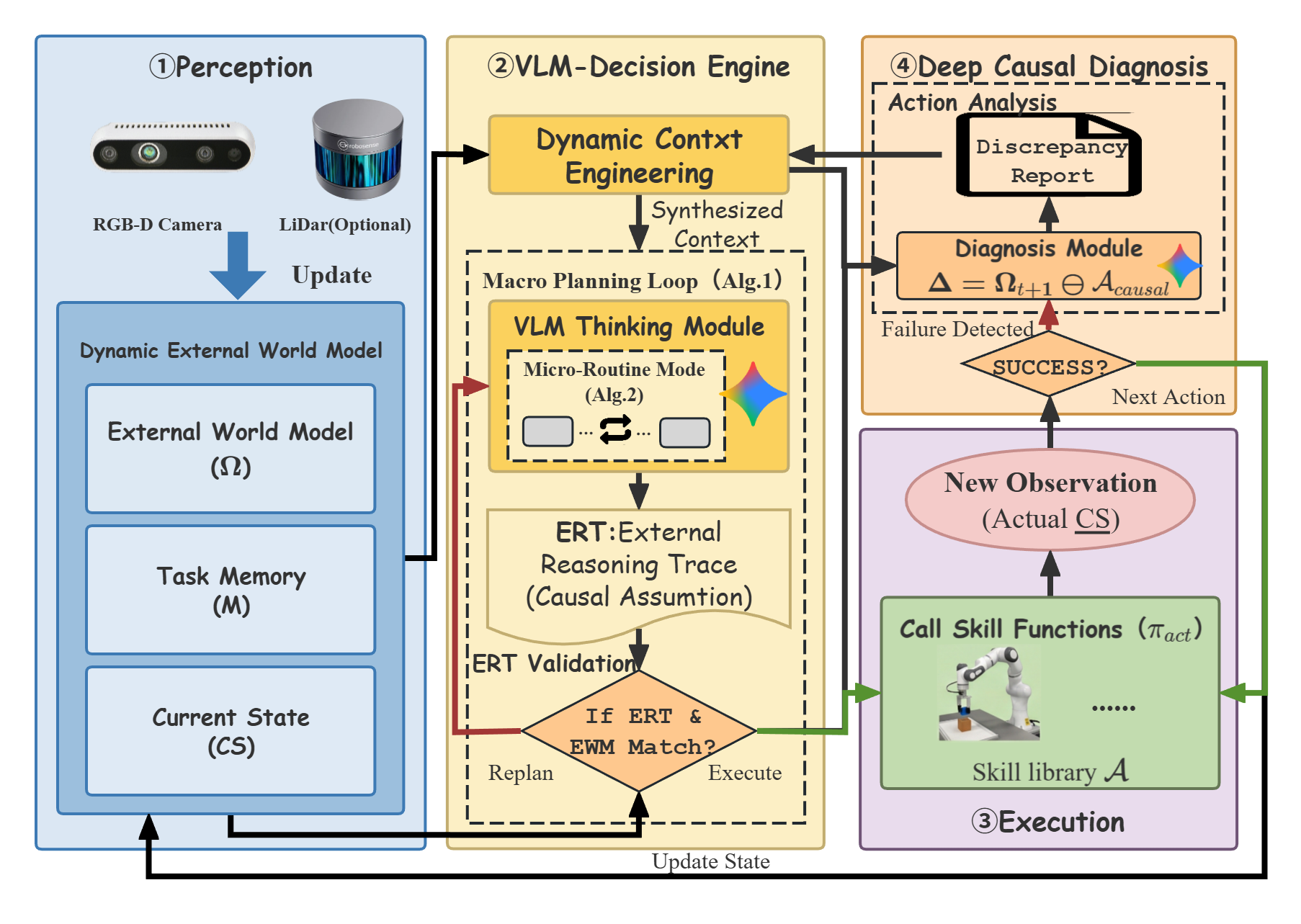} 
	\caption{\textbf{The Progressive VLM-Guided Planning and Deep Recovery Loop (Sec. 3.3).} 
		Dynamic Context Engineering distills the DEWM state ($\Omega, M, CS$) into a task-relevant prompt. The VLM then generates an External Reasoning Trace (ERT) which is mathematically validated before execution. On failure, the Deep Causal Diagnosis module computes the belief-outcome discrepancy ($\Delta = \Omega_{t+1} \ominus \mathcal{A}_{causal}$) to generate targeted feedback for belief correction and informed recovery.}
	\label{fig:decision_loop}
\end{figure*}

Having established a dynamic world model ($\Omega$), we now detail the decision-making engine: the ``Progressive VLM-Guided Planning Algorithm ($\Pi$)''. 
As illustrated in Fig.~\ref{fig:decision_loop}, this engine operates as a closed-loop controller that tightly couples VLM reasoning with the structured DEWM. Assuming a primitive skill library $\mathcal{A}$ for low-level execution, the decision process follows a rigorous \textit{State-Input-Process-Output} cycle to generate verifiable, adaptive behavior.

\subsubsection{Internal Working Memory: Tracking Task and Ego-States}
\label{sec:working_memory}
Robust long-horizon autonomy requires tracking not just the external environment, but also the agent's internal progress. We introduce two dedicated working memory modules:

\paragraph{a. Task Memory ($M_t$)}
This module serves as the ``project manager'', tracking global operation progress. Formally, $M_t = \langle \mathcal{T}_{DAG}, ss_t, \mathcal{H}_{ash} \rangle$:
\begin{itemize}
	\item $\mathcal{T}_{DAG}$: The ``Task Topology Path'', a Directed Acyclic Graph (DAG) decomposed from user instructions via VLM-based task decomposition (prompt template in \ref{app:task_decomposition}).
	\item $ss_t$: A state pointer to the active node in $\mathcal{T}_{DAG}$ (Status: \texttt{Executing}/\texttt{Completed}).
	\item $\mathcal{H}_{ash}$: The ``Action Sequence History'', a structured log recording every executed ERT. This serves as the ``episodic memory'' enabling causal diagnosis.
\end{itemize}

\paragraph{b. Constraint State ($CS_t$)}
The Constraint State $CS_t$ is central to our diagnosis mechanism (Contribution~3). While $\mathcal{S}$ tracks geometric configurations, $CS_t$ anchors the robot's semantic interaction phase (e.g., $\langle \texttt{Holding}, \text{blue\_block\_1} \rangle$, 
or $\langle \texttt{Approaching}, \text{target\_zone} \rangle$). Updates to $CS_t$ are event-driven, occurring only when the \texttt{causal\_assumption} of a validated ERT is functionally fulfilled. This mechanism provides the micro-topological anchor required for precise failure localization: when an anomaly occurs, $CS_t$ identifies \textit{which interaction phase} failed, enabling targeted diagnosis rather than generic task-level error handling.

\subsubsection{Dynamic Context Engineering}
\label{sec:context_engineering}
VLMs are inherently context-agnostic. To ground their reasoning and prevent ``lost-in-the-middle'' hallucinations, we design a ``Subgraph Extraction Pipeline'' that acts as a deterministic filter between the DEWM and the VLM.

Instead of feeding the entire world state $\Omega$, which introduces irrelevant noise, the mechanism utilizes the current subtask $ss_t$ and topological state $CS_t$ to index the Hierarchical Semantic Graph ($\mathcal{G}$). It extracts a ``Task-Relevant Subgraph''—the minimal set of objects and relations required for the current decision—and serializes it into a concise natural language prompt. This strategy minimizes token usage while maximizing information density. The subgraph extraction algorithm is detailed in \ref{app:subgraph_extraction}, with context-specific prompt templates for task decomposition (\ref{app:task_decomposition}), motion planning (\ref{app:coarse_planning}), and fine manipulation (\ref{app:fine_manipulation}).

\subsubsection{The Cognitive Core: External Reasoning Trace (ERT)}
\label{sec:ert}
To transform the VLM from an opaque ``black box'' into a verifiable agent, we enforce a structured output format called the ``External Reasoning Trace (ERT)''.
Every planning step produces a JSON object containing four key fields:
\begin{equation}
	ERT = \{ \pi_{act}, \mathcal{B}_{world}, \mathcal{A}_{causal}, \kappa_{conf} \}
\end{equation}
Here, $\pi_{act}$ is the action proposal, $\mathcal{B}_{world}$ represents structured assertions about the pre-condition, $\mathcal{A}_{causal}$ encodes the expected post-conditions, and $\kappa_{conf}$ is the self-evaluated confidence.

Crucially, the ERT functions as a probabilistic hypothesis subject to verification. Before execution, it undergoes a Multi-Layered Validation Protocol (see \ref{app:ert_validation}), which rigorously checks the consistency between the VLM's semantic beliefs ($\mathcal{B}_{world}$) and the geometric ground truth in DEWM. This mechanism acts as a cognitive firewall, intercepting hallucinations before they manifest as physical errors.

It is important to note that the ERT protocol focuses exclusively on Semantic Task Feasibility (e.g., verifying geometric containment). We explicitly decouple low-level kinematics and dynamics—such as collision avoidance and singularity handling—to the downstream execution layer. Exceptions at this lower level are captured as feedback signals and routed to the Constraint State ($CS$) module to trigger the diagnosis pipeline.

\subsubsection{Main Planning Loop and Adaptive Control}
\label{sec:planning_loop}

The core control logic is encapsulated in the progressive planning algorithm (Algorithm~\ref{alg:main_loop}). To ensure robustness, all inference calls are encapsulated within a ``Verify-Retry'' wrapper, \texttt{GetValidERT}.
The system operates in two hierarchical modes:
\begin{enumerate}
	\item Macro-Planning (Algorithm~\ref{alg:main_loop}): Handles high-level state transitions and discrete single-step actions.
	\item Micro-Routine (Algorithm~\ref{alg:micro_routine}): Synthesizes detailed primitive sequences for contact-rich manipulation.
\end{enumerate}

\begin{algorithm}
	\caption{Progressive VLM-Guided Planning (Main Loop)}
	\label{alg:main_loop}
	\small
	\begin{algorithmic}[1]
		\Require Task $\mathcal{T}$, Skill Library $\mathcal{A}$, Max Retries $N_{max}$
		\Ensure Task Status: \texttt{Completed} or \texttt{Failed}
		\State $\Omega, M, CS \leftarrow \Call{InitializeSystem}{\mathcal{T}}$
		\While{$\Call{GetStatus}{M} \neq \texttt{Completed}$}
		\State $\Omega_t, M_t, CS_t \leftarrow \Call{PerceiveAndUpdate}{\Omega_{t-1}}$
		\State $ss_t \leftarrow \Call{GetCurrentSubtask}{M_t}$
		
		\If{$ss_t$ is \texttt{ManipulationTask}}
		\State $status \leftarrow \Call{ExecuteMicroRoutine}{\Omega_t, M_t, CS_t, ss_t}$
		\Else
		\State $Prompt \leftarrow \Call{SynthesizeContext}{\Omega_t, M_t, CS_t}$
		\State $ERT, valid \leftarrow \Call{GetValidERT}{Prompt, N_{max}}$
		\If{$\neg valid$} \Return \texttt{Failed} \EndIf
		
		\State $skill \leftarrow \Call{ExtractAction}{ERT}$
		\State $status \leftarrow \Call{Execute}{skill, \mathcal{A}}$
		\EndIf
		
		\State $\Call{UpdateState}{M_t, CS_t, status, ERT}$
		\If{$status == \texttt{Failed}$}
		\State $\Call{DeepCausalDiagnosis}{\Omega_t, M_t, ERT}$ \Comment{Trigger Recovery}
		\EndIf
		\EndWhile
		\State \Return \texttt{Completed}
	\end{algorithmic}
\end{algorithm}

\begin{algorithm}
	\caption{Micro-Manipulation Subroutine}
	\label{alg:micro_routine}
	\small
	\begin{algorithmic}[1]
		\Require World $\Omega_t$, Memory $M_t$, State $CS_t$, Subtask $ss_t$
		\State $Prompt \leftarrow \Call{SynthesizeMicroPrompt}{\Omega_t, M_t, CS_t, ss_t}$
		\State $ERT_{micro}, valid \leftarrow \Call{GetValidERT}{Prompt, N_{max}}$
		\If{$\neg valid$} \Return \texttt{Failed} \EndIf
		
		\State $Seq \leftarrow \Call{ExtractSequence}{ERT_{micro}}$
		\For{$skill$ in $Seq$}
		\State $status \leftarrow \Call{Execute}{skill, \mathcal{A}}$
		\State $\Call{UpdateState}{M_t, CS_t, status, ERT_{micro}}$
		\If{$status == \texttt{Failed}$} \Return \texttt{Failed} \EndIf
		\EndFor
		\State \Return \texttt{Success}
	\end{algorithmic}
\end{algorithm}

\subsubsection{Deep Causal Diagnosis and Belief Correction ($CS$-Anchored Recovery)}
\label{sec:deep_diagnosis}

When a subtask fails (Line 18, Algorithm~\ref{alg:main_loop}), the recovery mechanism goes beyond simple retries. The Belief-Action History ($\mathcal{H}_{ash}$) retains the full ERT recorded prior to the failure. The \texttt{DeepCausalDiagnosis} function triggers a diagnostic flow by computing the semantic divergence $\Delta$ between the \texttt{causal\_assumption} in the ERT and the actual post-failure reality $\Omega_{t+1}$:
\begin{equation}
	\Delta = \Omega_{t+1} \ominus \mathcal{T}_{ert}.\mathcal{A}_{causal}
\end{equation}
Here, $\ominus$ denotes the set difference of semantic states (formally defined in \ref{app:discrepancy_operator}).

This calculation generates a precise \textbf{Belief-Outcome Discrepancy Report} (\ref{app:discrepancy_operator}), which is fed back to the VLM via \texttt{SynthesizeReplanPrompt} (template in \ref{app:failure_diagnosis}). This enables the agent to perform \textit{Belief Correction}—updating its internal model based on evidence—rather than merely generating a different plan, fundamentally distinguishing our approach from blind retry strategies.

\section{Experiments}
\label{sec:experiments}

To rigorously validate the effectiveness of the proposed VLM-DEWM cognitive architecture, we designed a comprehensive suite of experiments across high-fidelity simulation and real-world physical platforms. Our evaluation aims to answer three core Research Questions (RQs):
\begin{enumerate}
	\item Long-Horizon Reliability: Can DEWM provide persistent, cross-regional world memory to support state tracking and zero-shot spatial reasoning, outperforming pure VLMs that rely solely on context history?
	\item Precise Diagnosis: Compared to other memory-enhanced strategies, does the \textit{Micro-Topological State Anchoring} ($CS$) mechanism enable more accurate causal diagnosis and intelligent recovery in dynamic environments?
	\item Sim-to-Real Generalization: Can the core cognitive logic of VLM-DEWM generalize from simulation to physical interaction with real-world sensor noise and latency?
\end{enumerate}

\subsection{Experimental Setup}
\label{sec:setup}
This section details our experimental configuration. We describe the baseline methods in Section~\ref{sec:baselines}, platform implementation details in Section~\ref{sec:implementation}, experimental assumptions in Section~\ref{sec:assumptions}, task design in Section~\ref{sec:tasks}, and evaluation metrics in Section~\ref{sec:metrics}.

\subsubsection{Baseline Methods}
\label{sec:baselines}
We compare VLM-DEWM against two categories of baselines representing mainstream embodied planning paradigms.

\paragraph{a. Context-Only VLM Planners}
This category relies exclusively on the VLM's context window to maintain world understanding, without explicit external memory structures. We include both SOTA closed-source models (Gemini 2.5 Pro, GPT-5) and open-source models (Qwen3-VL-32B-Instruct, GLM-4.5V). For fairness, all VLM baselines are integrated into the same perception-planning loop, with their ``memory'' consisting solely of historical images and text prompts.

\paragraph{b. Memory-Enhanced Strategies}
This category constructs external memories for VLMs and is most relevant to our work. We select ``SAGE'' \cite{ref-sage} (a 3D Scene Graph approach) and ``GEAR'' \cite{ref-gear} (a SOTA VLM-RAG method) as representative competitors, reproducing them according to their official specifications.

\paragraph{c. Ours (VLM-DEWM)}
We utilize Gemini 2.5 Pro as the core ``cognitive collaborator''. This choice allows for a direct ``apples-to-apples'' comparison with the Pure Gemini 2.5 Pro baseline, isolating the performance gains attributable to the DEWM architecture rather than the underlying model capability.

\subsubsection{Implementation Details}
\label{sec:implementation}

\paragraph{a. Simulation Environment}
Our simulation platform is built on PyBullet and Gazebo. To rigorously isolate cognitive performance from perceptual noise, we adopt a "privileged information" strategy in simulation. Perfect segmentation masks (via \texttt{p.getCameraImage}) and ground-truth 6D poses are fed directly to the cognitive module. This ensures that the evaluation strictly benchmarks the DEWM's logic (memory, planning, diagnosis) rather than the upstream perception model errors.

\paragraph{b. Real-World Platform and Protocols}
The physical validation was conducted on a high-fidelity robotic cell designed to emulate industrial assembly (Fig.~\ref{fig:task3_real}).
\begin{itemize}
	\item Hardware Specs:
	We use a Franka Emika Panda (7-DoF, 1kHz control loop) equipped with a Robotiq 2F-85 gripper. Perception is provided by an Intel RealSense D435i (848$\times$480 resolution @ 30fps), mounted extrinsically.
	\item Sample Size ($N$):
	For Task 3, we conducted a total of 60 trials among which 20 trials were subjected to induced failures 
	(covering 3 failure modes: Part Slip, Obstacle, Target Moved).
	\item Success Criteria:
	A trial is successful if and only if the final assembly is completed within a 5mm position tolerance and 2 degree orientation tolerance relative to the CAD specification, verified by a calibrated overhead camera.
	\item Throughput Metric:
	We define \textit{Effective Throughput} as $\frac{3600s}{\text{Avg. Cycle Time} + \text{MTTR}}$. This penalizes methods that require manual reset after failure.
\end{itemize}

To filter perceptual noise, we adopt our prior Coarse-to-Fine 3D-Grounded Perception framework~\cite{ref-mypaper}, ensuring the evaluation focuses on cognitive performance rather than upstream perception errors. This framework integrates open-vocabulary 2D detection with 3D point cloud clustering and outlier removal. By enforcing both semantic consistency and geometric density constraints, it effectively filters out perceptual hallucinations before they enter the DEWM update loop. This allows our evaluation to focus specifically on the cognitive performance of the memory and reasoning modules.

\subsubsection{Experimental Assumptions and Priors}
\label{sec:assumptions}
To ensure rigor and alignment with target applications (e.g., smart manufacturing), we define the following assumptions:
\begin{itemize}
	\item Semi-Structured Environment: We assume object categories are known, but their poses, quantities, and topological relationships are unknown and dynamic.
	\item Shape Prior Library ($\mathcal{L}$): For high-precision interaction (Task 3), the system is provided with a Shape Prior Library containing CAD models and grasp priors, as described in Section~\ref{sec:perception_dynamics}.
	\item Fairness Note: This library $\mathcal{L}$ is provided to \textit{all} baselines (including SAGE and GEAR) to ensure the comparison focuses on ``memory and diagnosis'' rather than ``grasp pose learning''.
\end{itemize}

\subsubsection{Experimental Tasks}
\label{sec:tasks}
We designed three distinct tasks to evaluate the system's performance across long-horizon planning, large-scale memory retention, and dynamic resilience. To ensure relevance to RCIM readers, these tasks are modeled after core scenarios in flexible manufacturing and intralogistics.

\paragraph{Task 1: Long-Horizon Multi-Stage Assembly (Simulation)}
Modeled after a ``PCB Component Kitting and Insertion'' process. The robot operates in a multi-zone workcell ("Factories") containing various "components" (blocks/cylinders representing capacitors/chips).
\textit{Goal:} The VLM must parse a high-level manufacturing order (e.g., "Assemble the Red Kit and transfer to Zone C"), decomposing it into a sequence of 10+ manipulation steps (pick, place, stack), requiring strict adherence to geometric constraints and logical dependencies.

\paragraph{Task 2: Large-Scale Facility Inventory (Simulation)}
Simulates an automated inventory management task in a large-scale warehouse or town environment.
\textit{Goal:} The robot performs a random-walk exploration to build a semantic map of the facility. The system is then subjected to a rigorous "Inventory Audit" (Q\&A phase), assessing its ability to recall the location ("Where is the emergency stop?"), count ("How many pallets?"), and spatial relations of entities across the entire facility, testing the capacity of DEWM to handle large-scale static data.

\paragraph{Task 3: Dynamic Exception Handling (Real-World)}
Replicates a ``High-Mix Low-Volume Assembly Line'' prone to runtime disturbances.
\textit{Goal:} The robot performs a standard pick-and-place assembly on the physical Franka platform. We introduce real-world anomalies common in factory settings, such as part slippage (gripper failure) and unexpected obstacles (safety stops). The system must diagnose these interruptions in real-time and generate "Informed Recovery" strategies (e.g., "Re-grasp" or "Clear Obstacle") to minimize downtime, rather than blindly halting or retrying.

\subsubsection{Evaluation Metrics}
\label{sec:metrics}
Let $N$ be the total number of trials and $\mathbb{I}(\cdot)$ be the indicator function. We employ five quantitative metrics to evaluate system performance:

\paragraph{a. Industrial Reliability (Task Success Rate - TSR)}
Serves as the primary measure of end-to-end operational effectiveness (proxy for manufacturing yield rate). A trial is deemed successful only if all physical sub-goals are met and (for Task 2) all verification questions are answered correctly.
\begin{equation}
	TSR = \frac{1}{N} \sum_{i=1}^{N} \mathbb{I}(\text{Trial}_i = \text{Success}) \times 100\%
\end{equation}
where $N$ is the total number of experimental trials.

\paragraph{b. Digital Twin Fidelity (State Tracking Accuracy - STA)}
Quantifies the accuracy of the DEWM in maintaining object states when they are Out-of-View (OOV), acting as a metric for process monitoring quality. At each timestep $t$, we compare the estimated state $S_{EWM}(o)$ against the simulation ground truth $S_{GT}(o)$, averaged over the task duration $T$:
\begin{equation}
	STA = \frac{1}{T} \sum_{t=1}^{T} \frac{\sum_{o \in \mathcal{O}_{t, OOV}} \mathbb{I}(S_{EWM}(o) = S_{GT}(o))}{|\mathcal{O}_{t, OOV}|} \times 100\%
\end{equation}
This metric rigorously evaluates the "Object Permanence" capability required for reliable inventory tracking.

\paragraph{c. Operational Cost \& Latency (Planning Efficiency - PE)}
Acts as a proxy for economic cost (Token consumption) and cycle time (Inference latency). It is defined as the total count of VLM API calls per trial ($N_{calls}$), separating initial planning ($N_{init}$) from error-induced replanning ($N_{replan}$):
\begin{equation}
	PE = N_{calls} = N_{init} + N_{replan}
\end{equation}
A lower PE indicates a stable memory context that minimizes expensive model queries, directly translating to a more cost-effective production cycle.

\paragraph{d. Memory Storage Efficiency (Information Efficiency - IE)}
Measures the storage density of the system, critical for long-term deployment. It is calculated as the ratio of valid information queries answered ($N_{Info}$) to the total number of memory entries ($N_{Items}$) stored in the database:
\begin{equation}
	IE = \frac{N_{Info}}{N_{Items}} \times 100\%
\end{equation}
High IE indicates that the system stores highly relevant, structured knowledge rather than redundant raw logs (as seen in baseline SAGE).

\paragraph{e. Diagnostic Precision (Causal Diagnosis Accuracy - CDA)}
Evaluates the system resilience and ability to minimize ``Mean Time to Recovery (MTTR)''. It measures the percentage of diagnostic attempts where the system-generated reason $R_{gen}$ matches the ground truth reason $R_{GT}$:
\begin{equation}
	CDA = \frac{1}{M} \sum_{j=1}^{M} \mathbb{I}(R_{gen, j} = R_{GT, j}) \times 100\%
\end{equation}
where $M$ is the total number of diagnostic attempts (not failure events), as a single failure may require multiple diagnosis rounds before successful recovery. Unlike ``Blind Retry'' strategies (CDA=0), a high CDA enables ``Informed Replanning'', significantly reducing downtime.

\textit{Ground Truth Labeling Protocol:}
To ensure objectivity and reproducibility, $R_{GT}$ is determined through a post-hoc sensor log analysis protocol rather than subjective judgment. Specifically, for each induced failure event, we record synchronized multi-modal logs including: (1) gripper width and force-torque readings from the Franka controller, (2) RGB-D frames at 10 Hz, and (3) joint states and end-effector poses. 

\subsection{Main Results: Comparison with Baselines}
\label{sec:main_results}

We conducted comprehensive comparisons against the baselines defined in Section~\ref{sec:baselines} across both simulation and real-world platforms. The results for the three core tasks are presented in Tables~\ref{tab:task1_results}, \ref{tab:task2_results}, and \ref{tab:task3_results}.

\subsubsection{Task 1: Long-Horizon Assembly \& Reasoning (Simulation)}
\begin{figure}[H]
	\centering
	\includegraphics[width=0.4\linewidth]{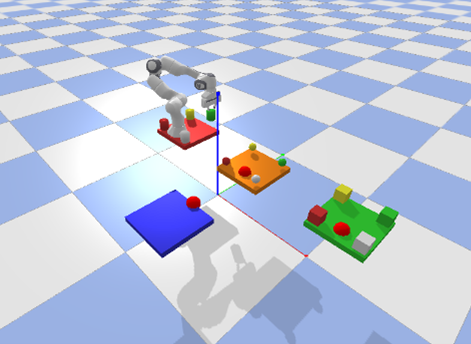}
	\caption{\textbf{Task 1 Environment Setup.} The multi-zone factory simulation scene containing blocks, spheres, and cylinders distributed across three distinct platforms. This raw visual input triggers the memory construction process.}
	\label{fig:task1_scene}
\end{figure}

Task 1 evaluates the system's ability to handle multi-step physical interactions and maintain state persistence. The scene comprises three distinct zones (``Factories'') containing blocks, spheres, and cylinders (see Fig.~\ref{fig:task1_scene}).

\begin{figure}[htbp]
	\centering
	\begin{subfigure}[b]{0.25\linewidth}
		\centering
		\includegraphics[width=\linewidth]{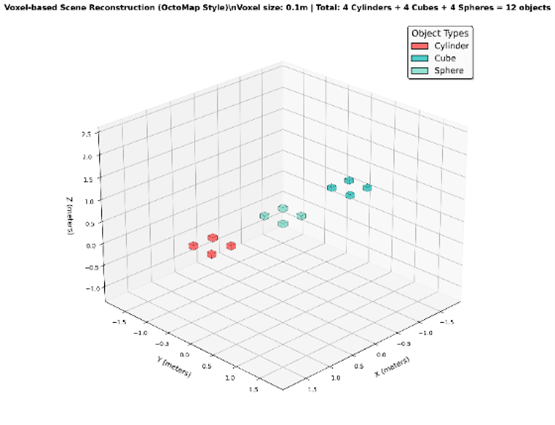}
		\caption{Background ($\mathcal{S}_{bg}$)}
	\end{subfigure}
	\hspace{0pt}
	\begin{subfigure}[b]{0.25\linewidth}
		\centering
		\includegraphics[width=\linewidth]{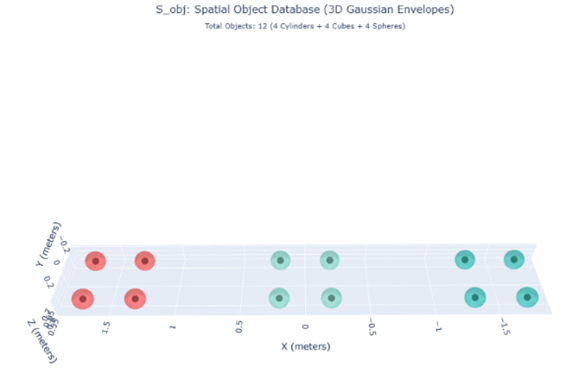}
		\caption{Object Registry ($\mathcal{S}_{obj}$)}
	\end{subfigure}
	
	\vspace{5pt}
	
	\begin{subfigure}[b]{0.5\linewidth}
		\centering
		\includegraphics[width=\linewidth]{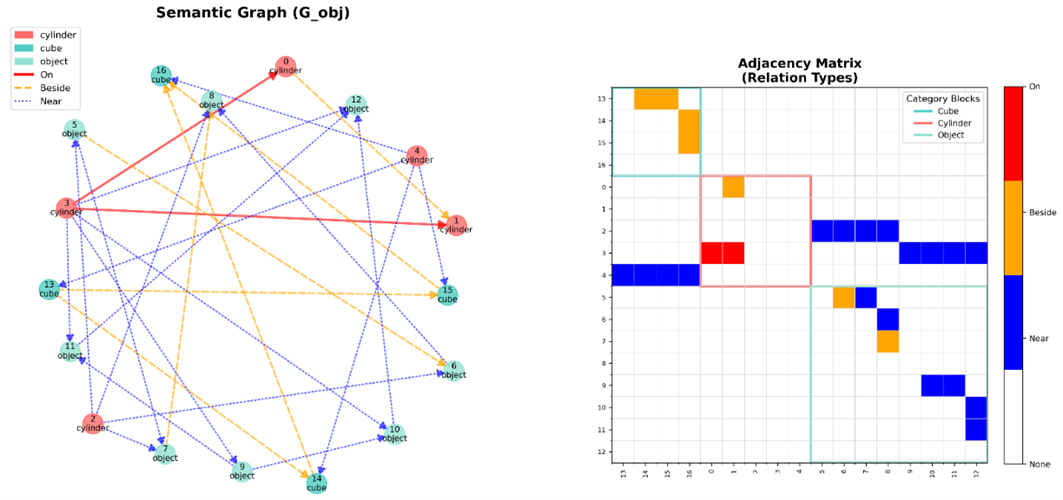}
		\caption{Initial Semantic Graph ($\mathcal{G}$) \& Adjacency Matrix}
		\label{fig:t1_g_init}
	\end{subfigure}
	
	\caption{\textbf{Initial Memory Construction in Task 1.} 
		\textbf{(a-b)} The Geometric Layer of DEWM separates the static background (Left) from dynamic objects (Right).
		\textbf{(c)} The Semantic Layer initializes the topological graph and its matrix representation, anchoring the scene's logical structure before manipulation.}
	\label{fig:task1_memory}
\end{figure}

\paragraph{a. Memory Construction Analysis}
Starting from the raw simulation input (Fig.~\ref{fig:task1_scene}), the system successfully decouples the static background $\mathcal{S}_{bg}$ from dynamic entities $\mathcal{S}_{obj}$ (Fig.~\ref{fig:task1_memory}a-b).
Simultaneously, the initial topological state is grounded in the Semantic Graph $\mathcal{G}$ (Fig.~\ref{fig:task1_memory}c).

\begin{figure}[htbp]
	\centering
	\begin{subfigure}[b]{0.6\linewidth}
		\centering
		\includegraphics[width=\linewidth]{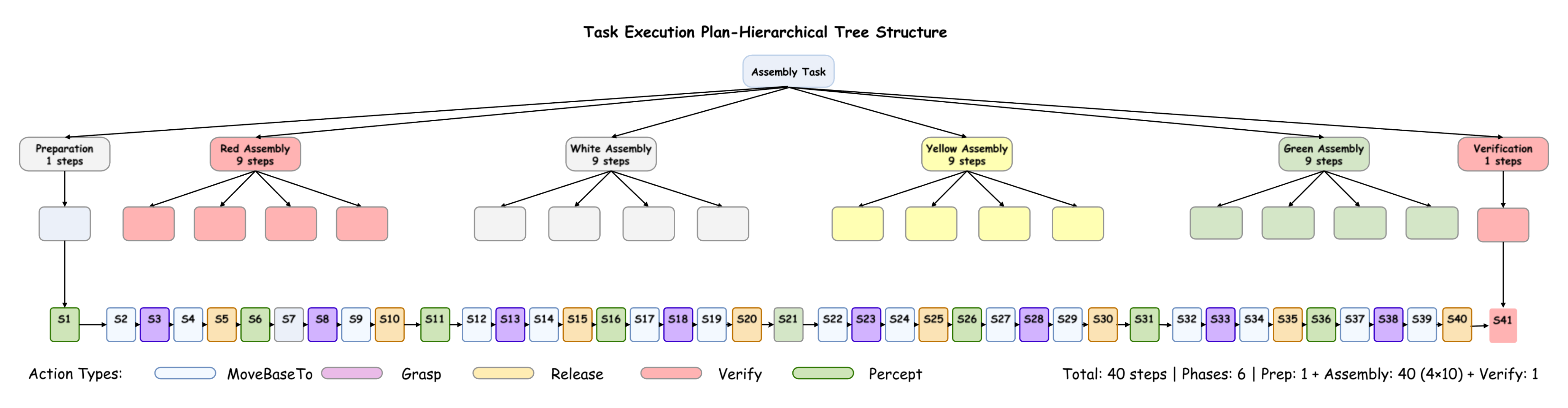}
		\caption{VLM-Generated Task Topology Path ($\mathcal{M}$)}
		\label{fig:t1_plan}
	\end{subfigure}
	
	\vspace{10pt}
	
	\begin{subfigure}[b]{0.6\linewidth}
		\centering
		\includegraphics[width=\linewidth]{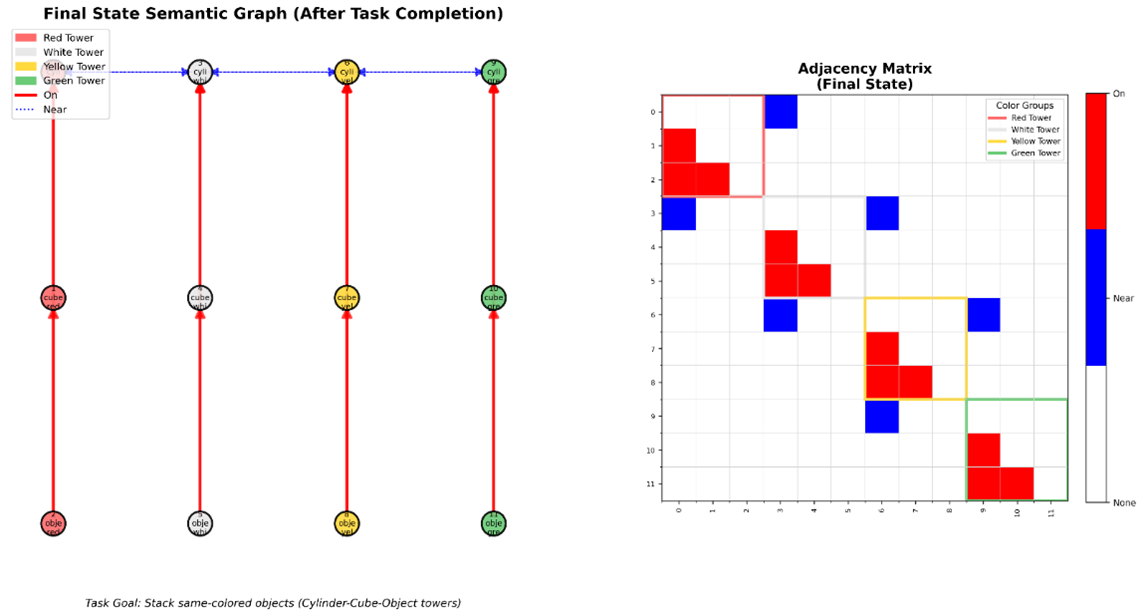}
		\caption{Post-Execution Dynamic Update}
		\label{fig:t1_update}
	\end{subfigure}
	
	\caption{\textbf{Reasoning and Dynamic Update Process.} 
		\textbf{(a)} Based on the initial memory, the VLM decomposes the instruction ``Complete Red Assembly'' into a hierarchical execution tree.
		\textbf{(b)} After execution, the DEWM reflects the new physical reality. By comparing this with Fig.~\ref{fig:t1_g_init}, we observe the real-time topology update (e.g., edge \texttt{On(cube, table)} is replaced by \texttt{On(cube, cylinder)}), verifying the system's dynamic tracking capability.}
	\label{fig:task1_reasoning}
\end{figure}

\paragraph{b. Reasoning and Execution Analysis}
The reasoning process is captured in Fig.~\ref{fig:task1_reasoning}. Specifically, the robot received the high-level instruction: ``Transport the red cube from Zone A and stack it onto the blue cylinder in Zone B.'' The Task Memory $\mathcal{M}$ (Fig.~\ref{fig:task1_reasoning}a) decomposes the high-level instruction into a structural plan. 
Most importantly, by comparing the initial matrix in Fig.~\ref{fig:task1_memory}(c) with the post-execution matrix in Fig.~\ref{fig:task1_reasoning}(b), we observe the dynamic graph update: relations are rewritten in real-time upon action completion (e.g., \texttt{On(cube, table)} $\to$ \texttt{On(cube, cylinder)}). This capability explains our high STA scores compared to the static memory of GEAR.

\paragraph{c. Quantitative Analysis}
As shown in Table~\ref{tab:task1_results}, VLM-DEWM outperforms all baselines ($N=50$ trials).
\begin{itemize}
	\item Ours: We achieved a 94.0\% TSR with near-perfect 100\% STA. The PE (75) is higher than pure VLMs because our architecture actively invokes the VLM for verification and geometric adjudication (Sec.~\ref{sec:ert}), whereas pure VLMs rely on a single, hallucination-prone generation (PE=1). Crucially, our 100\% IE confirms that the structured memory ($\mathcal{G}, \mathcal{S}$) stores zero redundant information.
	\item Memory-Enhanced Baselines: SAGE achieved a moderate 72\% TSR but suffered from severe inefficiency (PE=261, IE=12.26\%). Its unstructured ``agent-to-agent'' dialogue logging forces the VLM to parse redundant text, degrading performance. GEAR failed completely (0\% TSR) because its static RAG knowledge base could not update object locations after manipulation, leading to fatal conflicts between memory and reality.
	\item Pure VLMs: Despite receiving initial positions, pure VLMs struggled with the long horizon, exhibiting severe hallucination (TSR 0--28\%). Only the strongest models (Gemini 2.5 Pro, GPT-5) managed sporadic successes.
\end{itemize}

\begin{table}[htbp]
	\centering
	\caption{Quantitative Results for Task 1 (Long-Horizon Assembly)}
	\label{tab:task1_results}
	\small 
	\setlength{\tabcolsep}{5pt} 
	\begin{tabular}{@{}llcccc@{}} 
		\toprule
		\textbf{Method} & \textbf{VLM Core} & \textbf{TSR (\%)} $\uparrow$ & \textbf{STA (\%)} $\uparrow$ & \textbf{PE} $\downarrow$ & \textbf{IE (\%)} $\uparrow$ \\ 
		\midrule
		\textbf{VLM-DEWM (Ours)} & Gemini 2.5 Pro & \textbf{94.00} & \textbf{100.00} & 75 & \textbf{100.00} \\ 
		SAGE \cite{ref-sage} & Gemini 2.5 Pro & 72.00 & 83.33 & 261 & 12.26 \\ 
		GEAR \cite{ref-gear} & Gemini 2.5 Pro & 0.00 & -- & 148 & -- \\ 
		\midrule
		Gemini 2.5 Pro & Gemini 2.5 Pro & 28.00 & 39.74 & \textbf{1} & -- \\ 
		GPT-5 & GPT-5 & 12.00 & 34.62 & \textbf{1} & -- \\ 
		Qwen3-VL & Qwen3-VL & 0.00 & 30.77 & \textbf{1} & -- \\ 
		GLM-4.5V & GLM-4.5V & 0.00 & 15.38 & \textbf{1} & -- \\ 
		\bottomrule
	\end{tabular}
	
	\medskip
	\footnotesize\textit{Note:} Pure VLMs achieve PE=1 due to a single long-context prompt, but fail to complete the task (Low TSR).
\end{table}

\subsubsection{Task 2: Large-Scale Exploration (Simulation)}
This task evaluates memory construction and zero-shot spatial reasoning in a town-scale environment. We introduce ``Query Success Rate (QSR)'' as the primary metric, assessed via a graded question set (Table~\ref{tab:question_set}).

\begin{figure}[htbp]
	\centering
	\includegraphics[width=0.8\linewidth]{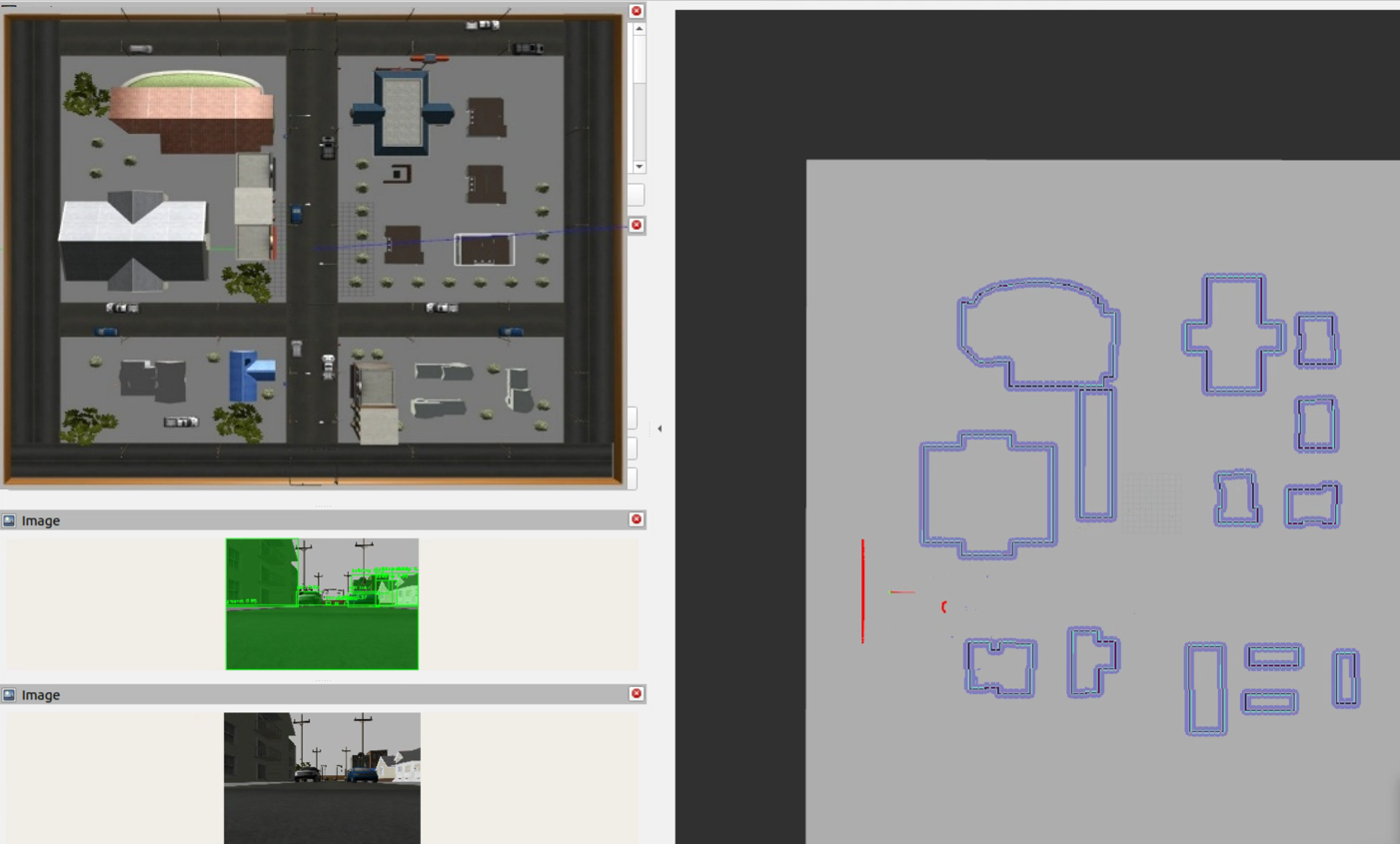} 
	\caption{\textbf{Task 2 Environment Setup.} Multi-modal inputs (LIDAR + Camera) + semantic segmentation acquired during the town-scale exploration. This raw sensory stream serves as the input for constructing the DEWM.}
	\label{fig:task2_scene}
\end{figure}

\begin{table}[htbp]
	\centering
	\caption{Task 2 Spatial Question Set Distribution}
	\label{tab:question_set}
	\small
	\begin{tabular}{lcl}
		\toprule
		\textbf{Difficulty} & \textbf{Count} & \textbf{Description} \\ 
		\midrule
		Easy & 15 & Basic Recognition (e.g., ``Is there a bank?'') \\ 
		Medium & 20 & Spatial Relations (e.g., ``Is the bench near the park?'') \\ 
		Hard & 5 & Multi-step Reasoning (e.g., Navigation heuristics) \\ 
		High & 10 & Daily Scenario Q\&A \\ 
		\bottomrule
	\end{tabular}
\end{table}

\begin{figure}[htbp]
	\centering
	\begin{subfigure}[b]{0.25\linewidth}
		\centering
		\includegraphics[width=\linewidth]{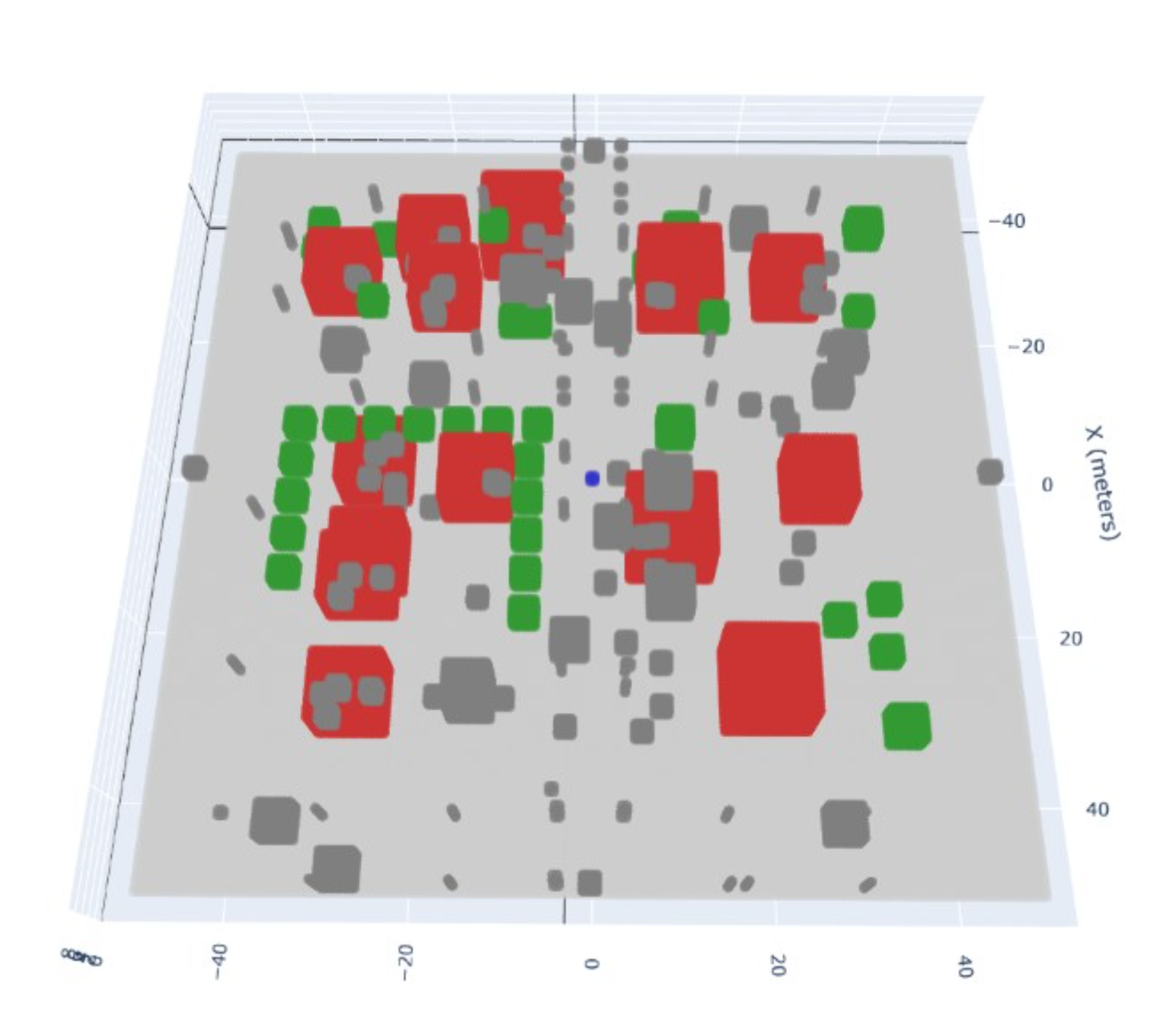}
		\hfill
		\caption{Global Map ($\mathcal{S}_{bg}$)}
		\label{fig:t2_sbg}
	\end{subfigure}
	\hspace{0pt}
	\begin{subfigure}[b]{0.25\linewidth}
		\centering
		\includegraphics[width=\linewidth]{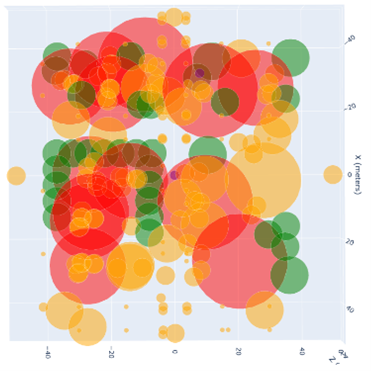}
		\hfill
		\caption{Landmarks ($\mathcal{S}_{obj}$)}
		\label{fig:t2_sobj}
	\end{subfigure}
	\vspace{5pt}
	\begin{subfigure}[b]{0.5\linewidth}
		\centering
		\includegraphics[width=\linewidth]{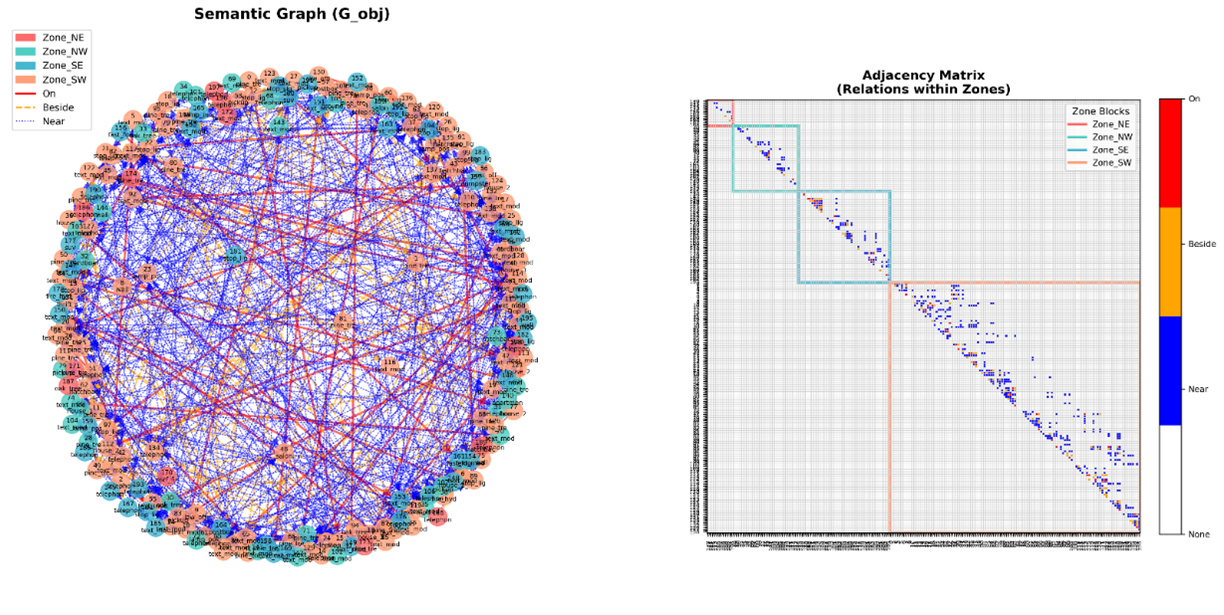}
		\caption{Semantic Graph Topology ($\mathcal{G}$)}
		\label{fig:t2_g}
	\end{subfigure}
	\caption{\textbf{Hierarchical Memory Construction in Task 2.} 
		\textbf{(a)} The reconstructed global semantic map serves as the metric substrate. 
		\textbf{(b)} Key landmarks (e.g., hydrants) are extracted as object-centric instances. 
		\textbf{(c)} The Semantic Graph abstracts these entities into topological relations (e.g., \texttt{Near(bench, park)}), enabling efficient spatial querying.}
	\label{fig:task2_memory}
\end{figure}

\paragraph{a. Hierarchical Memory Construction}
The DEWM construction process is visualized in Fig.~\ref{fig:task2_memory}.
Starting from raw inputs, the system builds a persistent global map $\mathcal{S}_{bg}$ and extracts key landmarks into the object registry $\mathcal{S}_{obj}$ (Fig.~\ref{fig:task2_memory}a-b). These geometric entities are abstracted into the town-scale semantic graph $\mathcal{G}$ (Fig.~\ref{fig:task2_memory}c), which serves as the knowledge base for answering the Q\&A set defined in Table~\ref{tab:question_set}.

\begin{figure}[htbp]
	\centering
	\includegraphics[width=0.6\linewidth]{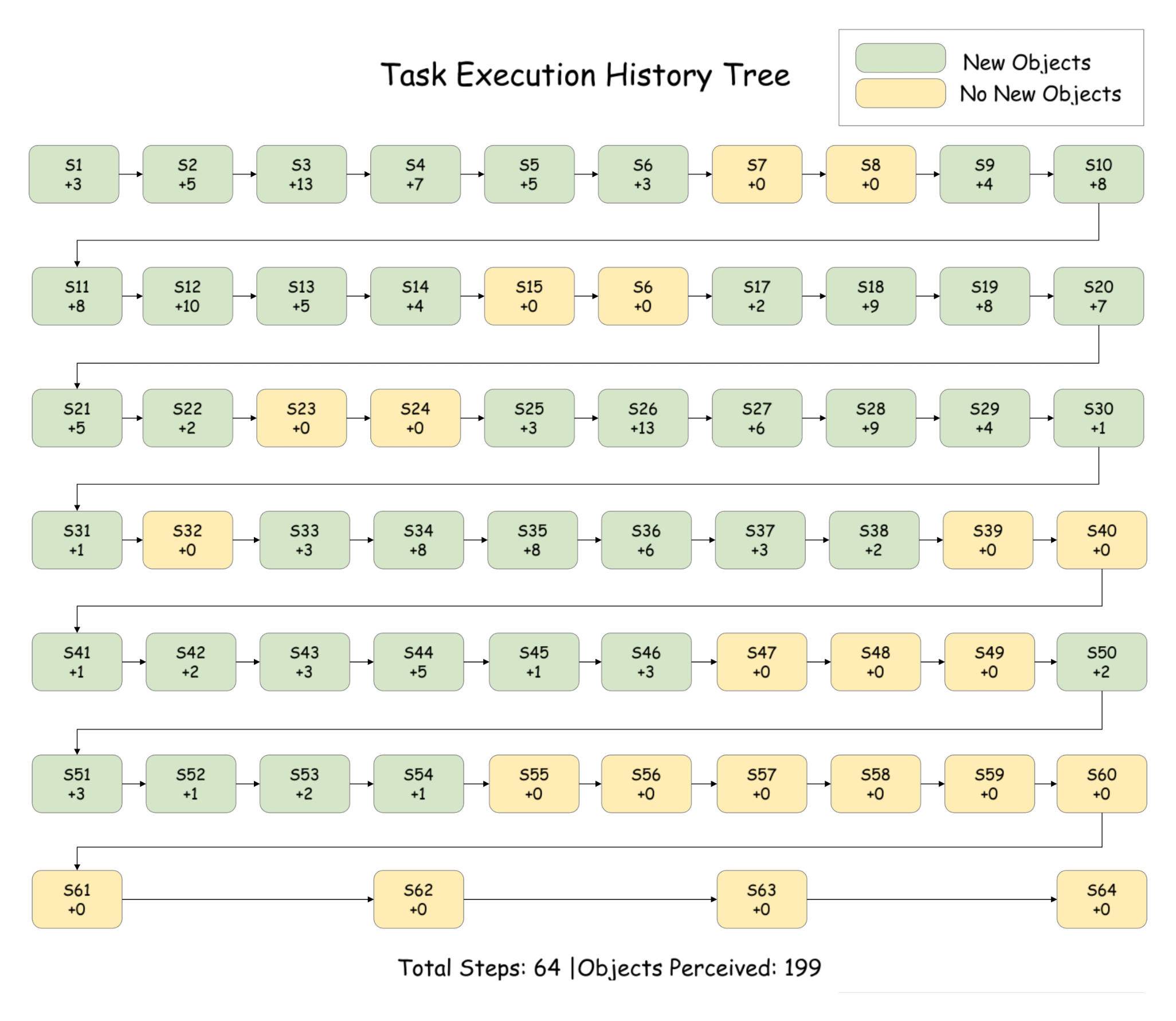}
	\caption{\textbf{Exploration Execution History ($\mathcal{M}$).} 
		The temporal tree visualizes the robot's 240-step exploration path. Color-coded nodes indicate memory events: \textbf{Green} nodes mark steps where new entities were discovered and registered into $\mathcal{G}$; \textbf{Yellow} nodes indicate steps with only re-observations, demonstrating the system's efficiency in avoiding redundant updates.}
	\label{fig:task2_process}
\end{figure}

\paragraph{b. Exploration Process Analysis}
In this scenario, the robot performed a 240-step random-walk exploration mission to map the unknown facility.
The temporal evolution of this mission is captured in the exploration history (Fig.~\ref{fig:task2_process}). The sparsity of Green nodes (New Entity) versus Yellow nodes (Re-observation) in the tree visually demonstrates our high Information Efficiency (IE), as the system effectively recognizes previously seen areas without redundant memory writes, creating a compact yet comprehensive map.

\paragraph{c. Quantitative Analysis}
As shown in Table~\ref{tab:task2_results}, VLM-DEWM exhibits overwhelming advantages in memory quality.
\begin{itemize}
	\item \textbf{Ours:} We achieved \textbf{94.0\% QSR} and \textbf{90.14\% IE}. By structuring the town into a hierarchical graph $\mathcal{G}$, our system converts spatial reasoning into efficient \textit{database queries} rather than text processing. The PE (509) reflects the necessary cost of building a persistent map over 240 observation points.
	\item \textbf{Baselines:} SAGE showed disastrous efficiency (PE=2162, IE=1.30\%). Its unstructured logging forced the VLM to perform ``needle-in-a-haystack'' retrieval on redundant text, limiting QSR to 52\%. Pure VLMs acted as streaming processors with no long-term retention (STA $\approx$ 29\%), failing to answer cross-regional queries.
\end{itemize}

\begin{table}[H]
	\centering
	\caption{Quantitative Results for Task 2 (Exploration \& Q\&A)}
	\label{tab:task2_results}
	\small 
	\setlength{\tabcolsep}{4pt} 
	\begin{tabular}{llcccc}
		\toprule
		\textbf{Method} & \textbf{VLM Core} & \textbf{QSR (\%)} $\uparrow$ & \textbf{STA (\%)} $\uparrow$ & \textbf{PE} $\downarrow$ & \textbf{IE (\%)} $\uparrow$ \\ 
		\midrule
		\textbf{VLM-DEWM (Ours)} & Gemini 2.5 Pro & \textbf{94.00} & \textbf{92.68} & 509 & \textbf{90.14} \\ 
		SAGE & Gemini 2.5 Pro & 52.00 & 56.10 & 2162 & 1.30 \\ 
		GEAR & Gemini 2.5 Pro & 74.00 & - & 240 & - \\ 
		\midrule
		Gemini 2.5 Pro & Gemini 2.5 Pro & - & 29.27 & 240 & - \\ 
		\bottomrule
	\end{tabular}
\end{table}

\subsubsection{Task 3: Real-World Dynamic Recovery}
\begin{figure}[htbp]
	\centering
	\includegraphics[width=0.9\linewidth]{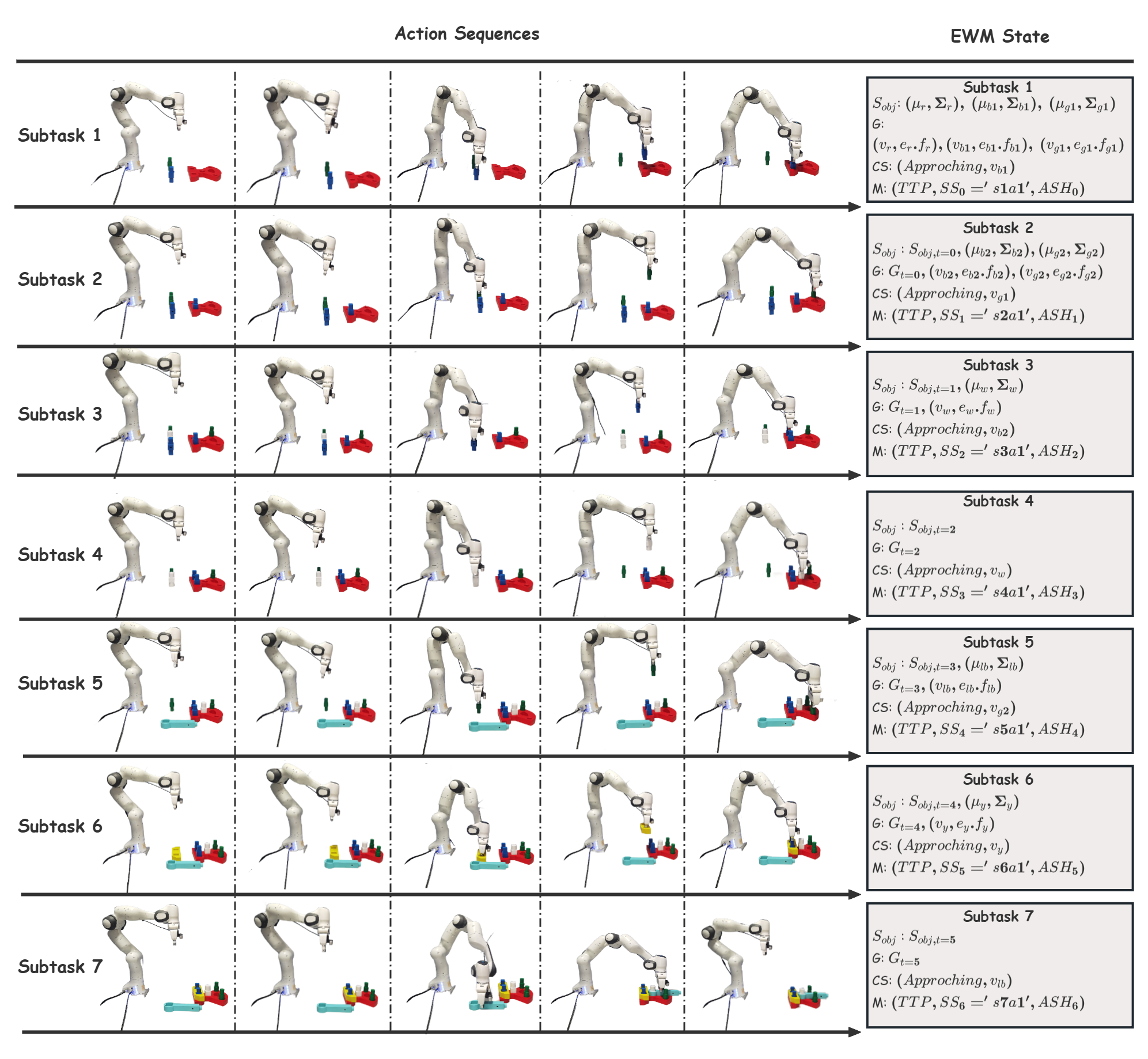}
	\caption{\textbf{Real-World Deployment.} The Franka Emika Panda performing the dynamic assembly task. The system successfully diagnoses and recovers from induced execution failures using the $CS$-based causal diagnosis.}
	\label{fig:task3_real}
\end{figure}

Finally, we validated the system on the physical Franka platform (Fig.~\ref{fig:task3_real}) using a dynamic assembly task with induced failures. 

Fairness Note: As specified in Section~\ref{sec:assumptions}, all methods utilized the same Shape Prior Library ($\mathcal{L}$) for grasp pose generation. This isolates the evaluation to strictly focus on \textit{memory, state tracking, and diagnosis}, decoupling it from low-level manipulation skill proficiency.

\paragraph{a. Mechanism and Qualitative Analysis}
Figure~\ref{fig:task3_real} illustrates the mechanism behind this resilience. 
In a standard cycle, the robot is tasked to retrieve a gear and insert it into a shaft. During the transport phase, we induced a ``Gripper Slip'' failure, causing the part to fall onto the table.

The core challenge in Task 3 is the dynamic evolution of the environment (e.g., new obstacles appearing during failure). Baselines like SAGE and GEAR rely on a static snapshot memory; once the environment changes, their knowledge base suffers from \textit{epistemic stagnation}, leading to inevitable planning conflicts.
In contrast, VLM-DEWM maintains a live topology. When the transport failure occurs (Fig.~\ref{fig:task3_real}, bottom), the system detects the topological breach (Expected: \texttt{Holding}, Actual: \texttt{Idle}). Instead of hallucinating success or freezing, it autonomously triggers a ``Re-perception and Search'' routine, effectively closing the cognitive loop in the physical world.

\paragraph{b. Quantitative Analysis}
Table~\ref{tab:task3_results} reveals a stark contrast: VLM-DEWM achieved an 95.0\% TSR, while all baselines failed completely (0\% TSR).
\begin{itemize}
	\item Cognitive Resilience via Iterative Diagnosis (Ours): A critical insight lies in the relationship between CDA (36.96\%) and TSR (95.0\%). Among the 20 induced failures, the system diagnosed the root cause \textit{immediately} in only 6 cases. However, in 11 cases, it successfully anchored the failure after multiple diagnostic rounds ($N < N_{max}=5$). This proves that VLM-DEWM possesses cognitive resilience: it does not require perfect zero-shot diagnosis but can progressively converge to the correct causal explanation through interaction, a capability entirely absent in standard VLM planners.
	\item Catastrophic Failure of Baselines: Baselines consistently scored 0\% CDA. Without the micro-topological anchoring provided by the $CS$ module, baselines suffered from ``state aliasing''—they could not distinguish between physically distinct failure modes (e.g., \texttt{Slip} vs. \texttt{Collision}). Consequently, they defaulted to a ``Blind Retry'' loop, repeating the same erroneous trajectory without updating their internal world model.
\end{itemize}

\begin{table}[H]
	\centering
	\caption{Real-World Dynamic Recovery Results ($N=60$)}
	\label{tab:task3_results}
	\small
	\begin{tabular}{llcc}
		\toprule
		\textbf{Method} & \textbf{VLM Core} & \textbf{TSR (\%)} $\uparrow$ & \textbf{CDA (\%)} $\uparrow$ \\ 
		\midrule
		\textbf{VLM-DEWM (Ours)} & Gemini 2.5 Pro & \textbf{95.00} & \textbf{36.96} \\ 
		SAGE & Gemini 2.5 Pro & 0.00 & 0.00 \\ 
		GEAR & Gemini 2.5 Pro & 0.00 & 0.00 \\ 
		Pure VLM & Gemini 2.5 Pro & 0.00 & 0.00 \\ 
		\bottomrule
	\end{tabular}
\end{table}

\subsubsection{Baseline Failure Analysis}
\label{sec:baseline_failure_analysis}

The zero TSR achieved by all baselines in Task 3 warrants detailed explanation, as it may raise concerns about task fairness. We emphasize that this outcome reflects \textit{fundamental architectural limitations} rather than unfair task design. Below we analyze the failure modes of each baseline category.

\paragraph{a. Pure VLM (Stateless Reasoning)}
Pure VLMs operate in a \textit{single-shot, stateless} manner: they receive the current image and task instruction, then generate an action without persistent memory. When a gripper slip occurs mid-transport, the VLM observes an image where the target object has ``disappeared'' from the gripper. Without access to the prior state (i.e., the object was being held), the VLM cannot distinguish between three possibilities: (1) the object was successfully placed, (2) the object was dropped, or (3) the object was never picked. In our experiments, Pure VLMs consistently misinterpreted slip events as 
successful placement, proceeding to the next subtask and causing cascading failures.

\paragraph{b. SAGE (Dialogue-Based Memory)}
SAGE~\cite{ref-sage} maintains memory through unstructured agent-to-agent dialogue logs. While this enables semantic persistence, it lacks two critical capabilities for failure recovery. First, SAGE has no \textit{geometric grounding}: when the robot reports ``I dropped the object,'' SAGE cannot determine \textit{where} the object fell because its memory contains only textual descriptions without metric coordinates. Second, SAGE lacks \textit{interaction-phase awareness}: it cannot distinguish whether a failure occurred during \texttt{Approaching}, \texttt{Grasping}, or \texttt{Transporting} phases. Without this micro-state anchoring (provided by $CS$ in our architecture), SAGE's recovery attempts 
were spatially ungrounded, causing the robot to search in incorrect locations.

\paragraph{c. GEAR (Static Knowledge Graph)}
GEAR~\cite{ref-gear} constructs a knowledge graph from initial observations but does not support \textit{dynamic relation updates}. In Task 3, when an object's spatial relation changes due to external disturbance (e.g., from \texttt{On(obj, gripper)} to \texttt{On(obj, floor)}), GEAR's static graph retains the outdated relation. This causes a fatal \textit{belief-reality 	divergence}: the planner continues to assume the object is held while perception shows otherwise. GEAR has no mechanism to reconcile this conflict, 
leading to repeated execution of invalid actions until timeout.

\paragraph{d. Architectural Implications}
These failure patterns validate our core thesis: failure recovery in dynamic environments requires (1) \textit{persistent geometric memory} to localize anomalies, (2) \textit{dynamic state updates} to track relation changes, and (3) \textit{interaction-phase anchoring} to diagnose failure causes. VLM-DEWM integrates all three through the $\mathcal{S}$-$\mathcal{G}$-$CS$ architecture, explaining its qualitative advantage in recovery scenarios. We acknowledge that Task 3 was designed to stress-test these capabilities; however, such disturbances (gripper slips, unexpected obstacles) are \textit{routine occurrences} in real manufacturing environments, making this evaluation industrially relevant rather than artificially adversarial.

\subsection{Ablation Studies}
\label{sec:ablation}

To rigorously validate the necessity of the four core components—$\mathcal{S}$ (Geometric Layer), $\mathcal{G}$ (Semantic Layer), $\mathcal{M}$ (Task Memory), and $CS$ (Constraint State)—we conducted a comprehensive ablation study on the real-world Franka platform.

\subsubsection{Experimental Setup}
We reused the task 3 (Dynamic Recovery) setup as it involves complex manipulation sequences that fully exercise all cognitive modules. We evaluated five model variants ($N=20$ trials each).

Protocol Note: To ensure independence, all scene-specific memories in DEWM were cleared between trials. However, aligned with industrial assembly scenarios, the Shape Prior Library ($\mathcal{L}$) containing grasp poses for known parts was retained.

Additional Metric: We introduced a specific Q\&A set for this ablation, comprising 5 spatial questions (e.g., ``Relation between Part A and B?'') and 5 semantic questions (e.g., ``Count of green parts?''), to measure the internal consistency of the memory.

\subsubsection{Model Variants}
\begin{itemize}
	\item \textbf{VLM-DEWM (Full):} The complete architecture proposed in this paper.
	\item \textbf{w/o $\mathcal{S}$ (No Geometric Anchor):} $\mathcal{S}$ is removed. The system retains $\mathcal{G}$, $\mathcal{M}$, and $CS$, but VLM reasoning lacks physical anchoring and the ``Geometric Adjudication'' capability.
	\item \textbf{w/o $\mathcal{G}$ (No Semantic Interface):} $\mathcal{G}$ is removed. The VLM must attempt to extract context directly from raw geometric data ($\mathcal{S}$), losing the structured symbolic interface.
	\item \textbf{w/o $\mathcal{M}$ (No Task Memory):} $\mathcal{M}$ is removed. The system possesses a world model ($\mathcal{G}+\mathcal{S}$) and micro-states ($CS$) but lacks the Task Topology Path (TTP) and Action State History (ASH).
	\item \textbf{w/o $CS$ (No Diagnostic Anchor):} The micro-topological state module is removed. In failure events, the system perceives only macro-task failure without micro-state granularity.
\end{itemize}

\begin{table}[t]
	\centering
	\caption{Ablation Study of Core Components in Real-World Task 3 ($N=20$)}
	\label{tab:ablation}
	\resizebox{\columnwidth}{!}{%
		\begin{tabular}{lccccc}
			\toprule
			\textbf{Method} & \textbf{TSR (\%)} $\uparrow$ & \textbf{CDA (\%)} $\uparrow$ & \textbf{STA (\%)} $\uparrow$ & \textbf{QSR (\%)} $\uparrow$ & \textbf{PE (calls)} $\downarrow$ \\ 
			\midrule
			\textbf{VLM-DEWM (Full)} & \textbf{95.00} & \textbf{68.97} & \textbf{100.00} & \textbf{100.00} & \textbf{54} \\ 
			w/o $\mathcal{S}$ & 0.00 & - & 0.00 & 60.00 & - \\ 
			w/o $\mathcal{G}$ & 0.00 & - & 0.00 & 0.00 & - \\ 
			w/o $\mathcal{M}$ & 90.00 & - & 100.00 & 100.00 & 266 \\ 
			w/o $CS$ & 55.00 & 0.00 & 97.50 & 95.50 & - \\ 
			\bottomrule
		\end{tabular}%
	}
\end{table}

\subsubsection{Analysis of Ablations}
As presented in Table~\ref{tab:ablation}, all four components proved indispensable.

\paragraph{a. VLM-DEWM (Full)}
The full architecture served as the benchmark, achieving a 95.0\% TSR and perfect state tracking (100\% STA/QSR). 
\textit{Failure Analysis:} Only one trial failed completely, caused by a hardware safety stop triggered by LIDAR glare, which positioned the target out of the workspace. This hardware-level exception confirms that the cognitive logic itself remained robust, achieving a high diagnosis rate (68.97\% CDA) for induced failures.

\paragraph{b. w/o $\mathcal{S}$: Epistemic Detachment}
This variant failed completely (0\% TSR). When an object was deliberately moved, the semantic belief in $\mathcal{G}$ (e.g., \texttt{On(blue\_cube, table)}) became obsolete. Lacking $\mathcal{S}$ for ``Geometric Adjudication'' (Section 3.3.3) and spatial memory for Out-of-View objects, the system could not detect this deviation.
\textit{Result:} The VLM generated plans based on stale beliefs, causing execution at incorrect locations. While it could answer semantic questions (QSR 60\%), the 0\% STA reveals a fatal detachment between ``Belief'' and ``Reality''.

\paragraph{c. w/o $\mathcal{G}$: Symbol Grounding Failure}
Removing $\mathcal{G}$ rendered the system inoperable (0\% TSR). $\mathcal{G}$ acts as the ``Translation Interface'' that bi-directionally binds high-level VLM symbols to physical entities in $\mathcal{S}$. Without this interface, the VLM could not ``ground'' its instructions into raw pose data, failing to generate even the first step. Consequently, semantic awareness vanished (STA/QSR $\approx$ 0\%), proving $\mathcal{G}$ is critical for VLM-compatible interaction.

\paragraph{d. w/o $\mathcal{M}$: Efficiency Collapse (Transient Memory)}
This variant highlights the efficiency role of $\mathcal{M}$. Without TTP and ASH, the system exhibited ``transient memory'', forgetting long-term goals after every step. It was forced to replan from scratch continuously.
\textit{Result:} While perception modules ($\mathcal{G}, \mathcal{S}, CS$) kept accuracy high (90\% TSR), the computational cost exploded. In trials with induced failures, the redundant replanning drove the VLM API calls to 266 (compared to 54 for Full), confirming $\mathcal{M}$'s role in guaranteeing long-horizon efficiency.

\paragraph{e. w/o $CS$: Diagnostic Blindness}
This variant validated the ``Deep Causal Diagnosis'' mechanism (Section 3.4.4). In the 20 trials, we observed 6 gripper slips and 3 perception deviations. Lacking the micro-topological states (e.g., \texttt{Approaching} vs. \texttt{Transporting}) as anchors, the system perceived these only as generic macro-failures (CDA 0\%). Lacking a detailed Discrepancy Report, it defaulted to ``Blind Retry'', which fails in dynamic environments. The significant drop in TSR (to 55\%) confirms that $CS$ is essential for intelligent recovery.

\paragraph{f. Remark on ERT}
Notably, our ablation does not include a ``w/o ERT'' variant. In our architecture, ERT is not a removable component but the fundamental communication protocol (Section 3.4.2) connecting ``Thought'' (VLM) and ``Memory'' (DEWM). As shown by the ``Pure VLM'' baselines in Section 4.1.1 (TSR $\approx$ 0\%), removing this structured verification protocol causes the system to fail fundamentally.

\subsection{Discussion}
\label{sec:discussion}

Our experiments in simulation and real-world environments (Sections~\ref{sec:main_results}, \ref{sec:ablation}) fundamentally validate the efficacy of our core contribution: the ERT cognitive Paradigm. The results unambiguously demonstrate that the failure of baseline methods stems from their opaque nature—they lack a verifiable ``transaction'' mechanism (ERT) and a persistent ``database'' (DEWM).

Conversely, the success of VLM-DEWM derives from this verifiable closed loop. In this section, we deeply analyze how the specific data structures of DEWM ($\mathcal{G}, \mathcal{S}, \mathcal{M}, CS$) serve as indispensable components for this paradigm (Section~\ref{sec:discuss_memory}) and candidly discuss the system's failure boundaries (Section~\ref{sec:limitations}).

\subsubsection{Structured Memory Beyond Context Windows}
\label{sec:discuss_memory}

The ablation results (Table~\ref{tab:ablation}) reveal a clear architectural necessity: 
each DEWM component addresses a specific cognitive deficit that causes baseline failures. 
The Semantic Graph ($\mathcal{G}$) provides a structured, queryable interface that eliminates 
the inefficient ``needle-in-a-haystack'' retrieval seen in SAGE (IE=1.30\%). 
The Spatial Network ($\mathcal{S}$) serves as a geometric anchor that maintains 
belief-reality synchronization, preventing the ``epistemic detachment'' observed in 
GEAR and Pure VLMs. 
The Task Memory ($\mathcal{M}$) provides temporal scaffolding for long-horizon planning, 
avoiding the efficiency collapse (PE=266) of memoryless agents. 
Finally, the Constraint State ($CS$) enables micro-topological diagnosis, 
transforming ``blind retry'' into ``informed recovery''. 
The complete failure of single-component ablations (TSR=0\% for w/o $\mathcal{G}$ and 
w/o $\mathcal{S}$) confirms that these modules form an indivisible cognitive whole.

\subsubsection{Limitations and Failure Mode Analysis}
\label{sec:limitations}
While VLM-DEWM demonstrates robust cognitive resilience, it operates under specific assumptions. Violating these boundaries leads to performance degradation or failure.

\paragraph{a. World Model Pollution (Upstream Perception Errors)}
The quality of DEWM is bounded by the Symbolic Perception Interface (Section 3.3.1). The system cannot self-correct corrupted sensory inputs; it faithfully ``memorizes'' errors.
\begin{itemize}
	\item \textit{Scenario:} If the segmentation model erroneously merges two red cubes into one large cuboid, $\mathcal{G}$ will register a single entity. A VLM instruction to ``Pick one of the cubes'' becomes unexecutable.
	\item \textit{Scenario:} As observed in one failed trial, LIDAR glare or transparent objects can prevent $\mathcal{S}$ from building stable Gaussian envelopes, causing ``Geometric Adjudication'' to return null results continuously.
\end{itemize}

\paragraph{b. Open-Vocabulary Disaster (Closed-World Assumption)}
Our architecture, specifically the category fields in $\mathcal{G}$ and the Shape Prior Library $\mathcal{L}$, implicitly assumes a semi-structured world where object categories are known.
\begin{itemize}
	\item \textit{Scenario:} If a novel object (e.g., a ``yellow banana'') appears on the industrial assembly table, VLM-driven Data Association (Section 3.3.2) will fail. The system may either ignore it as an obstacle or, worse, hallucinate it as a geometrically similar known part (e.g., ``yellow cylinder'') and attempt a forceful insertion, leading to physical damage.
\end{itemize}

\paragraph{c. Scope of Physical Verification (Dynamics Gap)}
Our current "Physical Consistency Check" focuses on geometric feasibility (e.g., collisions, containment). 
We explicitly acknowledge that the current framework does not verify dynamics (e.g., payload stability under acceleration) or kinematics (e.g., joint limits during trajectory). 
In our implementation, these checks are offloaded to the low-level impedance controller. Future work will integrate a physics simulator (e.g., MuJoCo) into the ERT verification loop to catch dynamic failures \textit{ex-ante}.

\paragraph{d. Granularity Limits of Causal Diagnosis}
The $CS$ module enables topological diagnosis but lacks physical granularity.
\begin{itemize}
	\item \textit{Scenario:} If an object slips during transport due to low friction, the system correctly diagnoses the topological mismatch (Expected: \texttt{Holding}, Actual: \texttt{Idle}). However, it does not know the \textit{physical cause} (friction vs. force). Consequently, the recovery strategy (``Retry Grasp'') might fail again if the underlying physical parameter (grasp force) is not adjusted.
\end{itemize}

\paragraph{e. The ``Cognitive Cost'' of Deliberation}
Our architecture is deliberative, not reactive. The ``Verify-Reflect'' loop imposes a latency cost.
\begin{itemize}
	\item \textit{Scenario:} In highly dynamic tasks requiring reflex-level responses (e.g., catching a rolling ball), the system's cognitive cycle time (reflected in PE) creates a ``Reactivity Gap''. By the time an ERT is generated and verified, the world state may have already drifted, rendering the action obsolete.
\end{itemize}

\section{Conclusion and Future Work}
\label{sec:conclusion}

This paper addresses two critical challenges in VLM-based long-horizon robotic planning: world-state drift caused by implicit, transient state encoding, and opaque failure diagnosis that necessitates costly blind retries. We attribute these failures to a fundamental responsibility mismatch—asking VLMs to simultaneously perform semantic reasoning and maintain dynamic world states within a limited context window. To resolve this, we proposed VLM-DEWM, an architecture that decouples persistent state maintenance from semantic reasoning, thereby transforming the VLM into a verifiable planning component.

Our experiments support the efficacy of this architectural separation. The DEWM framework, through its dual geometric ($\mathcal{S}$) and semantic ($\mathcal{G}$) grounding, ensures persistent state tracking across stations; ablation studies confirm that removing either component consistently fails. Furthermore, by structuring VLM outputs into Externalizable Reasoning Traces (ERT), the system enables pre-execution validation against physical consistency, effectively converting opaque decisions into auditable transactions. This structural transparency also facilitates interaction-phase tracking ($CS$), which proved essential for targeted failure localization and informed recovery. Quantitatively, the approach significantly improved state-tracking accuracy and recovery success rates while reducing inference overhead compared to memory-augmented VLM baselines.

Beyond specific performance gains, this work highlights a broader design principle: the interface between foundation models and physical systems should be structured, verifiable, and auditable. The ERT protocol instantiates this by exposing VLM reasoning to external validation, a strategy likely to be critical for deploying foundation models in safety-critical industrial settings where accountability and traceability are required.

Future work will focus on relaxing the current assumptions of the framework. First, we aim to incorporate uncertainty explicitly by modeling pose covariance within the world model, enabling risk-aware planning. Second, we plan to extend the system to open-vocabulary scenarios, moving beyond the current semi-closed Shape Library to handle novel object categories dynamically. Finally, fusing force-torque and acoustic signals into the constraint state ($CS$) offers a promising avenue for finer-grained diagnosis, allowing the system to distinguish between physically distinct failure modes such as slip and collision.

In summary, VLM-DEWM provides a persistent, interpretable memory core that supports cross-station consistency and resilient recovery. This transition from reactive intelligence to deliberative autonomy offers a principled path for embodied agents to operate reliably in dynamic physical environments.

\section*{CRediT authorship contribution statement}

\begin{itemize}
	\item \textbf{Guoqin Tang}:  Conceptualization, Methodology, Validation, Formal analysis, Investigation, Writing–original draft, Software, Data curation, Visualization.
	
	\item \textbf{Qingxuan Jia}: Resources, Project administration, Funding acquisition. 
	
	\item \textbf{Gang Chen}:  Conceptualization, Investigation, Writing– review \& editing, Funding acquisition.
	
	\item \textbf{Tong Li}:  Conceptualization, Investigation, Writing– review \& editing, Funding acquisition.
	
	\item \textbf{Zeyuan Huang}: Writing – review \& editing, Validation, Formal analysis, Conceptualization.
	
	\item \textbf{Zihang Lv}: Investigation, Validation.
	
	\item \textbf{Ning Ji}: Investigation, Funding acquisition.
	
\end{itemize}

\section*{Declaration of competing interest}
The authors declare that they have no known competing financial interests or personal relationships that could have appeared to influence the work reported in this paper.

\section*{Data availability}
The authors do not have permission to share data.

\section*{Acknowledgments}
This work is supported by the National Natural Science Foundation of China under Grant No. 62573064; and State Key Laboratory of Robotics and Systems (HIT) (SKLRS-2025-KF-07). All authors have read and agreed to the published version of the manuscript.

\appendix
\section{Implementation Details of the Symbolic Perception Interface}
\label{app:perception_pipeline}

This appendix details the technical implementation of the symbolic perception pipeline described in Section~\ref{sec:perception_dynamics}. The objective is to transform raw RGB-D observations $O_t$ into a structured symbolic snapshot $Env_t$ efficiently.

\subsection{Vocabulary-Guided 2D Attention Focusing}
\label{app:A1}
To avoid computationally expensive dense reconstruction of irrelevant background, we leverage the VLM to filter the scene in the 2D domain first.

\paragraph{a. Task-Relevant Category Generation:}
Given the task description $\mathcal{T}$ and RGB image $I_t$, the VLM infers a list of relevant object categories $\mathcal{C}_{task}$:
\begin{equation}
	\mathcal{C}_{task} = \{ c_1, c_2, \dots \} = \text{VLM}_{inference}(I_t, \mathcal{T})
\end{equation}
\textit{Example:} For $\mathcal{T}=$ "Fetch the drink", $\mathcal{C}_{task} \rightarrow$ ['bottle', 'can', 'cup', 'table'].

\paragraph{b. Open-Vocabulary Segmentation:}
This list $\mathcal{C}_{task}$ serves as the text prompt for an open-vocabulary segmentation model (specifically, we use ``Mask2Former'' trained on cityscapes/COCO). The model outputs a set of binary masks $\mathcal{M}$:
\begin{equation}
	\mathcal{M}(u, v) = \{ (c_i, \text{ID}_i) \mid \text{SegModel}(I_t, \mathcal{C}_{task}) \}
\end{equation}
Pixels not belonging to $\mathcal{C}_{task}$ are masked out ($null$), significantly reducing the data volume for downstream 3D processing.

\subsection{Sparse 3D Lifting and Coordinate Transformation}
\label{app:A2}
We establish a precise mapping from selected 2D pixels to the robot's 3D workspace.

\paragraph{a. Depth Back-Projection:}
For a valid pixel $(u, v)$ with depth $d = D_t(u, v)$, we project it to the camera coordinate frame $p_c$:
\begin{equation}
	p_c = d \cdot \mathbf{K}^{-1} [u, v, 1]^\top
\end{equation}
where $\mathbf{K}$ is the camera intrinsic matrix.

\paragraph{b. Kinematic Transformation:}
The point is then transformed to the robot base frame $p_b$ using the kinematic chain:
\begin{equation}
	p_b = {^b\mathbf{T}_e} \cdot {^e\mathbf{T}_c} \cdot p_c
\end{equation}
where ${^b\mathbf{T}_e}$ is the end-effector pose (from forward kinematics) and ${^e\mathbf{T}_c}$ is the hand-eye calibration matrix.

\paragraph{c. Ray-Casting Correspondence:}
To robustly associate semantic masks with noisy depth data, we employ a ray-casting check. A ray $\mathbf{r}(t)$ originating from the optical center passes through pixel $(u, v)$ into the point cloud $P_b$. A point $p \in P_b$ is assigned the label of pixel $(u, v)$ if and only if it lies within a distance threshold $\epsilon$ of the ray:
\begin{equation}
	\frac{\| \mathbf{r}(t) \times (p - \mathbf{c}) \|}{\| \mathbf{r}(t) \|} < \epsilon
\end{equation}
This generates a sparse set of labeled 3D points: $Map_t = \{ (p_b^k, \text{label}_k, \text{ID}_k) \}$.

\subsection{Distance-Aware Multi-Resolution Quantization}
\label{app:A3}
We partition the workspace into concentric zones based on the robot's physical reach capabilities.

\paragraph{a. Threshold Definition:}
\begin{itemize}
	\item $r_{near}$ (e.g., 0.8m): The maximum reach of the robotic arm without base movement.
	\item $r_{far}$ (e.g., 2.0m): The trusted sensing range for reliable object identification.
\end{itemize}

\paragraph{b. Quantization Logic:}
Based on the centroid distance $d_i = \| \bar{p}_i \|$, the object's geometry $geom_i$ is constructed as:
\begin{itemize}
	\item \textbf{Near Zone ($d_i < r_{near}$):} We retain the raw point cloud $\mathcal{PC}_i$. This supports high-fidelity ICP registration for grasping.
	\item \textbf{Mid Zone ($r_{near} \le d_i < r_{far}$):} We discretize points into a Voxel Grid $\mathcal{V}_i$ (voxel size = 2cm). This preserves topology while compressing data for collision checking.
	\item \textbf{Far Zone ($d_i \ge r_{far}$):} We abstract the object as a ``Directed Billboard'' $\mathcal{B}_i = \langle I_{crop}, \vec{v}_{view}, \mathbf{p}_{cen} \rangle$, where $I_{crop}$ is the RGB crop, $\vec{v}_{view}$ is the viewing vector, and $\mathbf{p}_{cen}$ is the centroid. This acts as a visual proxy for situational awareness with minimal memory cost.
\end{itemize}

\subsection{Final Symbolic Object Encapsulation}
\label{app:A4}
Finally, we aggregate the fragmented mappings into coherent object instances. For each unique instance ID detected in the mask set, we gather its geometric primitives and query the VLM for a descriptive summary.
The final observation snapshot $Env_t$ is a set of tuples, where each object $obj_i^t$ is defined as:
\begin{equation}
	obj_i^t = \langle \text{id}_i, \text{label}_i, geom_i^t, \text{desc}_i \rangle
\end{equation}
\begin{itemize}
	\item $\text{id}_i$: Unique instance identifier (e.g., "blue\_block\_1").
	\item $\text{label}_i$: Semantic category from VLM (e.g., "block").
	\item $geom_i^t$: The heterogeneous geometric representation derived in \ref{app:A3}.
	\item $\text{desc}_i$: A VLM-generated attribute description (e.g., "A metallic cube, possibly heavy").
\end{itemize}

\subsection{Perception Pipeline Triggering Policy}
\label{app:perception_timing}

The perception-to-memory pipeline (Section~\ref{sec:perception_dynamics}) does not run at a fixed frame rate. Instead, we employ an event-driven triggering policy to balance computational cost and state freshness.

\subsubsection{Trigger Events}

The full perception pipeline is invoked upon:
\begin{enumerate}
	\item \textbf{Action Boundary:} Before and after each skill execution (e.g., pre-grasp, post-place).
	\item \textbf{Zone Transition:} When the robot enters a new $\mathcal{G}_{zone}$ region.
	\item \textbf{Anomaly Detection:} When F/T feedback or motion tracking indicates unexpected events.
	\item \textbf{Explicit Query:} When the VLM requests updated state for a specific object.
	\item \textbf{Timeout:} If no trigger occurs for $T_{max} = 10$ seconds (background refresh).
\end{enumerate}

\subsubsection{Handling Drift During Non-Observation Periods}

Between perception events, object states in $\mathcal{S}_{obj}$ are \textbf{not updated}. We address potential drift through:

\paragraph{a. Conservative Covariance Inflation:}
For each object $o_i$ not observed in the current cycle, we inflate 
its position uncertainty by a fixed process noise:
\begin{equation}
	\boldsymbol{\Sigma}_t(o_i) \gets \boldsymbol{\Sigma}_{t-1}(o_i) + \mathbf{Q}_{drift}
\end{equation}
where $\mathbf{Q}_{drift} = \text{diag}(0.001, 0.001, 0.001)\,\text{m}^2$ represents per-cycle drift uncertainty (approximately 3~cm standard deviation per axis). This accumulation signals to the planner that unobserved objects have increased positional uncertainty.

\paragraph{b. Pre-Action Verification:}
Before any manipulation action targeting object $o_i$, the system \textit{mandatorily} triggers a perception update for $o_i$, regardless of recency. This ensures no action is executed based on stale beliefs.

\paragraph{c. Static Object Assumption:}
For objects tagged as ``static'' in $\mathcal{G}$ (e.g., fixtures, tables), we assume zero drift and skip re-observation unless explicitly invalidated.

\subsubsection{Complexity Implications}

The event-driven policy means the per-cycle costs in \ref{app:complexity} are \textbf{amortized}, not per-frame. In our experiments:
\begin{itemize}
	\item Task 1 (Assembly): $\sim$15 perception events over $\sim$120 seconds ($\approx 0.125$ Hz effective rate)
	\item Task 2 (Exploration): $\sim$240 perception events over $\sim$600 seconds ($\approx 0.4$ Hz effective rate)
	\item Task 3 (Recovery): Variable, with bursts during failure diagnosis
\end{itemize}
This is significantly lower than continuous SLAM-style updates, validating the efficiency claims.

\section{Algorithmic Details for DEWM Dynamics and Maintenance}
\label{app:dewm_dynamics}

This appendix provides the specific hyperparameters, noise models, and prompt templates required to reproduce the DEWM dynamics.

\subsection{Data Association and Similarity Scoring}
\label{app:B1}

This section specifies the association cost, gating rules, and VLM prompt templates used to match observations $o_k \in Env_t$ to memory entries $s_i \in \mathcal{S}_{obj}$ in the Hungarian assignment.

\subsubsection{Association Cost and Weights}
We construct a hybrid cost matrix:
\begin{equation}
	C_{ki} = \lambda_{iou} \cdot (1 - \text{IoU}(o_k, s_i)) + \lambda_{sem} \cdot (1 - \text{sim}(o_k, s_i)),
	\quad \lambda_{iou} + \lambda_{sem} = 1 .
\end{equation}
Unless otherwise stated, we set $(\lambda_{iou}, \lambda_{sem})=(0.7, 0.3)$, prioritizing geometric consistency while allowing semantic cues to resolve visually similar instances.

\subsubsection{IoU Definition (3D/2D Fallback)}
For each observation--memory pair, we compute IoU over \textit{3D oriented bounding volumes} when depth is reliable. If depth is unreliable (e.g., reflective surfaces or missing depth), we fall back to 2D mask IoU in the image plane using the most recent view:
\begin{equation}
	\text{IoU}(o_k, s_i) =
	\begin{cases}
		\frac{\text{Vol}(\mathcal{B}_{o_k} \cap \mathcal{B}_{s_i})}{\text{Vol}(\mathcal{B}_{o_k} \cup \mathcal{B}_{s_i})}, & \text{3D OBBs available} \\
		\frac{\text{Area}(\mathcal{M}_{o_k} \cap \mathcal{M}_{s_i})}{\text{Area}(\mathcal{M}_{o_k} \cup \mathcal{M}_{s_i})}, & \text{2D fallback}
	\end{cases}
\end{equation}
where $\mathcal{B}$ denotes an oriented bounding volume and $\mathcal{M}$ denotes a 2D segmentation mask.

\subsubsection{Semantic Similarity Prompt and Parsing}
The semantic similarity $\text{sim}(o_k, s_i)\in[0,1]$ is queried from the VLM when geometric cues alone are ambiguous (e.g., multiple nearby candidates or visually similar parts). The inputs \texttt{\{loc\}} are 3D Euclidean coordinates relative to the robot base, and \texttt{\{feat\}} are short attribute strings extracted from perception.

\begin{tcolorbox}[colback=gray!5, colframe=black, title=Prompt Template: Semantic Similarity]
	\textbf{System:} You are a spatial reasoning assistant. \\
	\textbf{Context:} An object described as a `\{category\_i\}' was last seen at \{loc\_old\} with features \{feat\_i\}. A new object is perceived at \{loc\_new\} with features \{feat\_new\}. \\
	\textbf{Query:} Is this the same object instance? Analyze the spatial offset and feature consistency. Answer with a single confidence score between 0.0 and 1.0.
\end{tcolorbox}

We use greedy decoding with temperature $0$ and parse the first numeric token in the VLM output as $\text{sim}(o_k, s_i)$. If parsing fails, we set $\text{sim}=0$ as a conservative default. We use $\tau_{sim}=0.5$ as the neutral decision boundary.

\subsubsection{Hungarian Assignment and Match Gating}
Given $C_{ki}$, we solve a one-to-one assignment using the Hungarian algorithm. A proposed match $(o_k, s_i)$ is accepted only if it passes a match-quality gate:
\begin{equation}
	C_{ki} \le \tau_{match}.
\end{equation}
Unless otherwise stated, we set $\tau_{match}=0.55$. This gate rejects low-confidence matches and prevents identity switches in cluttered scenes.

\subsubsection{Confirmation Window for Unmatched Observations}
If $o_k$ is not matched to any active entry (or all candidate matches violate $\tau_{match}$), it enters a confirmation window and is tracked tentatively for $n_{confirm}$ consecutive observations before being promoted to a new active entity. We set $n_{confirm}=2$ by default. Tentative tracks that are not re-observed within the window are discarded, suppressing one-off false detections and preventing ghost artifacts.

\subsection{Adaptive Fusion and VLM Gain Modulation}
\label{app:B2}

This section provides the implementation details for the VLM-modulated Gaussian envelope fusion described in Section~\ref{sec:adaptive_fusion}.

\subsubsection{Quantifying Observation Uncertainty ($\boldsymbol{\Sigma}_{obs}$)}

The observation covariance $\boldsymbol{\Sigma}_{obs}$ in Equations~(\ref{eq:sigma_fusion})--(\ref{eq:mu_fusion}) is computed based on the geometric abstraction available at time $t$:

\begin{itemize}
	\item \textbf{Near-Field (Point Cloud):} 
	\begin{equation}
		\boldsymbol{\Sigma}_{obs} = \frac{1}{N-1}\sum_{i=1}^{N}(\mathbf{p}_i - \boldsymbol{\mu}_{obs})(\mathbf{p}_i - \boldsymbol{\mu}_{obs})^\top + \epsilon \mathbf{I},
	\end{equation}
	where $N$ is the point count and $\epsilon = 10^{-6}\,\text{m}^2$ prevents singularities.
	
	\item \textbf{Mid/Far-Field (Bounding Volume):} 
	\begin{equation}
		\boldsymbol{\Sigma}_{obs} = \text{diag}(\sigma_x^2, \sigma_y^2, \sigma_z^2),
	\end{equation}
	with default $(\sigma_x, \sigma_y, \sigma_z) = (0.01, 0.01, 0.02)$ meters.
\end{itemize}

\subsubsection{Gain Mapping Table}

Table~\ref{tab:gamma_mapping} specifies the numerical mapping $\gamma(\alpha_t)$ used in all experiments.

\begin{table}[h]
	\centering
	\caption{Mapping from VLM Reliability Judgment to Fusion Weight}
	\label{tab:gamma_mapping}
	\begin{tabular}{lcc}
		\toprule
		\textbf{VLM Judgment} ($\alpha_t$) & \textbf{Weight} ($\gamma$) & \textbf{Effect} \\
		\midrule
		High & 1.0 & Full observation contribution \\
		Medium & 0.5 & Balanced fusion \\
		Low & 0.1 & Conservative update \\
		Bad & 0.01 & Near-freeze at prior \\
		\bottomrule
	\end{tabular}
\end{table}

\subsubsection{Prompt Template for Reliability Assessment}

We use greedy decoding ($T=0$) and parse the first matching token.

\begin{tcolorbox}[colback=gray!5, colframe=black, title=Prompt: Observation Reliability Assessment]
	\textbf{Context:} You are assessing the reliability of a sensor observation. \\
	\textbf{Object:} \{object\_category\} at position \{position\} \\
	\textbf{Observation Conditions:} \\
	- Occlusion level: \{occlusion\_percentage\}\% \\
	- Lighting: \{lighting\_quality\} \\
	- Object material: \{material\_type\} \\
	- Depth confidence: \{depth\_confidence\} \\
	\textbf{Task:} Rate the observation reliability as one of: High, Medium, Low, or Bad. \\
	Respond with only one word.
\end{tcolorbox}

\subsection{Scale-Aware Robust ICP Solver Details}
\label{app:B3}
To handle scale discrepancies (e.g., between canonical CAD models and real-world instances) and sensor noise, we solve the joint optimization problem:
\begin{equation}
	(\hat{\mathbf{T}}, \hat{s}) = \arg\min_{\mathbf{R}, \mathbf{t}, s} \sum_{i=1}^{N} \rho_\delta \left( \| s (\mathbf{R} p_i + \mathbf{t}) - q_i \|^2 \right)
\end{equation}
where $p_i \in \mathcal{P}^*$ are source model points and $q_i$ are corresponding target points.

\paragraph{a. Robust Cost Function:}
We use the Huber loss $\rho_\delta(\cdot)$ to mitigate the effect of outliers (e.g., segmentation artifacts):
\begin{equation}
	\rho_\delta(r) = 
	\begin{cases} 
		\frac{1}{2}r^2 & \text{for } |r| \le \delta \\
		\delta (|r| - \frac{1}{2}\delta) & \text{otherwise}
	\end{cases}
\end{equation}
We adaptively set $\delta = 2.5 \cdot \text{median}(r)$ at each iteration.

\paragraph{b. Closed-Form Update Steps:}
While the joint optimization is non-convex, we decouple it into two convex sub-problems solved iteratively.At each iteration, we compute weights from the Huber loss:
\begin{equation}
	w_i = \frac{\rho'_\delta(r_i)}{r_i}, \quad 
	r_i = \| s(\mathbf{R}p_i + \mathbf{t}) - q_i \|
\end{equation}
where $\rho'_\delta$ is the derivative of the Huber function ($\rho'_\delta(r) = \min(1, \delta/|r|) \cdot \text{sign}(r)$).

\begin{itemize}
	\item \textbf{Pose Update ($\mathbf{R}, \mathbf{t}$):} Using the weighted centroids $\bar{p} = \frac{\sum w_i p_i}{\sum w_i}$ and $\bar{q} = \frac{\sum w_i q_i}{\sum w_i}$, we compute the cross-covariance matrix $\mathbf{H} = \sum_{i} w_i (p_i - \bar{p})(q_i - \bar{q})^\top$. The rotation is derived via SVD:
	\begin{equation}
		\mathbf{U} \mathbf{\Lambda} \mathbf{V}^\top = \text{SVD}(\mathbf{H}), \quad \mathbf{R} = \mathbf{V} \mathbf{U}^\top
	\end{equation}
	Check: If $\det(\mathbf{R}) < 0$, we correct $\mathbf{V}$ by multiplying its last column by -1 to ensure a valid rotation.
	
	\item \textbf{Scale Update ($s$):} With $\mathbf{R}, \mathbf{t}$ fixed, the optimal scale is analytically derived:
	\begin{equation}
		s = \frac{\sum_{i} w_i (q_i - \bar{q})^\top \mathbf{R} (p_i - \bar{p})}{\sum_{i} w_i \| p_i - \bar{p} \|^2}
	\end{equation}
\end{itemize}

\subsection{Background Map Maintenance Implementation}
\label{app:B4}
The background model $\mathcal{S}_{bg}$ employs a hybrid volumetric-visual representation.

\paragraph{a. Probabilistic Volumetric Fusion:}
For voxel occupancy $O(n)$, we use the standard log-odds notation $L(n) = \log \frac{P(O(n))}{1-P(O(n))}$. The update rule for measurement $z_t$ is:
\begin{equation}
	L(n | z_{1:t}) = \text{clamp} \left( L(n | z_{1:t-1}) + L(n|z_t), 
	l_{min}, l_{max} \right)
\end{equation}
where $l_{min}=-2, l_{max}=3.5$ prevent over-confidence.

Simultaneously, we maintain a semantic distribution $P(C_k | z_{1:t})$ over classes $C_k \in \{\text{wall, table, floor, ...}\}$ using a recursive Bayesian update:
\begin{equation}
	P(C_k | z_{1:t}) = \eta \, P(z_t | C_k) \, P(C_k | z_{1:t-1})
\end{equation}
where $\eta = \left( \sum_{k'} P(z_t | C_{k'}) P(C_{k'} | z_{1:t-1}) \right)^{-1}$ ensures normalization.

\paragraph{b. Directed Billboard Management:}
For ray endpoints exceeding $r_{far}$, we generate a visual proxy. 
\begin{itemize}
	\item \textbf{Storage:} A billboard $\mathcal{B}_k = \langle I_{crop}, \vec{v}_{view}, \mathbf{p}_{cen} \rangle$ stores the RGB appearance $I_{crop}$ and the viewing direction $\vec{v}_{cam}$.
	\item \textbf{Retrieval Metric:} When the planner queries a frontier direction $\vec{v}_{query}$, we retrieve the most relevant billboard $k^*$ maximizing cosine similarity:
	\begin{equation}
		k^* = \arg\max_{k} \left( \frac{\vec{v}_{query} \cdot \vec{v}_{view}^k}{\| \vec{v}_{query} \| \| \vec{v}_{view}^k \|} \right)
	\end{equation}
	This simple metric effectively handles the view-dependent nature of open boundaries (e.g., looking "out" a window vs. looking "at" a wall).
\end{itemize}

\subsection{Statistical Basis for Geometric Verification}
\label{app:B5}
The verification module treats relation proposals as statistical hypothesis tests. We assume the position estimates of objects follow multivariate normal distributions $\mathcal{N}(\boldsymbol{\mu}, \boldsymbol{\Sigma})$.

\paragraph{a. Containment Test ($A \supset B$):}
The condition verifies if the bulk of $B$'s probability mass lies within the high-likelihood region of $A$. We compute the Mahalanobis distance of $B$'s centroid relative to $A$'s distribution:
\begin{equation}
	D^2_{contain} = (\boldsymbol{\mu}_B - \boldsymbol{\mu}_A)^\top \boldsymbol{\Sigma}_A^{-1} (\boldsymbol{\mu}_B - \boldsymbol{\mu}_A)
\end{equation}
Note that only $\boldsymbol{\Sigma}_A$ is used in the metric because the constraint is defined by the boundary of $A$.

\paragraph{b. Contact Test ($A \text{ touches } B$):}
This tests the hypothesis that the two distributions overlap significantly. The uncertainty of the relative difference vector $\Delta \mathbf{x} = \mathbf{x}_B - \mathbf{x}_A$ is given by the convolution of variances (sum of covariance matrices):
\begin{equation}
	D^2_{contact} = (\boldsymbol{\mu}_B - \boldsymbol{\mu}_A)^\top (\boldsymbol{\Sigma}_A + \boldsymbol{\Sigma}_B)^{-1} (\boldsymbol{\mu}_B - \boldsymbol{\mu}_A)
\end{equation}

\paragraph{c. Threshold Selection ($\chi^2_{thresh}$):}
The squared Mahalanobis distance follows a Chi-square distribution with $k$ degrees of freedom. Since we evaluate spatial relations in 3D translational space ($x, y, z$), we set $k=3$.
For a strict 95\% confidence level (i.e., we are 95\% sure the relation holds), we consult the $\chi^2$ table:
\begin{equation}
	P(X \le \chi^2_{thresh} | k=3) = 0.95 \implies \chi^2_{thresh} \approx 7.815
\end{equation}
Any relation with a distance score $D^2 > 7.815$ is rejected as a statistical anomaly or noise.

\subsection{Long-Term Memory Lifecycle Management}
\label{app:B6}
To ensure robustness in long-horizon operations, the ``Memory Curation'' is not merely a deletion process but a full Entity Lifecycle Management system. We define three states for every memory entity: \texttt{Active}, \texttt{Uncertain}, and \texttt{Archived}.

\paragraph{a. State Transitions and Decay Dynamics:}
When the robot performs a perception scan in Zone $Z_k$, the confidence $c_t$ of an unobserved entity $v_i$ follows an exponential decay model, representing the increasing probability that the object has been moved or removed:
\begin{equation}
	c_t(v_i) = c_{t-1}(v_i) \cdot \lambda_{decay}
\end{equation}
Entities transition states based on thresholds:
\begin{itemize}
	\item \textbf{Active $\to$ Uncertain:} When $c_t < \tau_{uncertain}$ (set to 0.6). The planner adds a "Verify" sub-task before interacting.
	\item \textbf{Uncertain $\to$ Archived:} When $c_t < \tau_{archive}$ (set to 0.1). The entity is removed from the spatial index tree (KD-Tree) to optimize query speed but retained in cold storage (Disk).
\end{itemize}

\paragraph{b. Restoration of Archived Entities (The "Zombie" Problem):}
A critical feature is the ability to recover from false negatives. If a new observation $o_{new}$ appears but fails to match any \texttt{Active} entity, the system queries the \texttt{Archived} database. If a match is found (using the logic in B.1), the entity is restored to \texttt{Active} state with $c_t$ reset to 1.0, preserving its historical semantic attributes (e.g., "fragile").

\paragraph{c. Curation Algorithm:}
The complete lifecycle maintenance logic is detailed in Algorithm~\ref{alg:curation}.

\begin{algorithm}
	\caption{Entity Lifecycle Maintenance}
	\label{alg:curation}
	\begin{algorithmic}[1]
		\Require DEWM $\Omega$, Current Zone $Z_k$, Observed IDs $\mathcal{O}_k$
		\State $\mathcal{K}_k \leftarrow \Omega.\text{GetEntitiesInZone}(Z_k)$
		\State $\mathcal{U}_k \leftarrow \mathcal{K}_k \setminus \mathcal{O}_k$ \Comment{Unobserved entities}
		
		\State \textit{// Phase 1: Decay and Archive}
		\For{$id$ in $\mathcal{U}_k$}
		\State $v \leftarrow \Omega.\text{GetEntity}(id)$
		\State $v.confidence \leftarrow v.confidence \times \lambda_{decay}$
		\If{$v.confidence < \tau_{archive}$}
		\State $\Omega.\text{ArchiveEntity}(id)$ \Comment{Move to cold storage}
		\ElsIf{$v.confidence < \tau_{uncertain}$}
		\State $v.state \leftarrow \texttt{Uncertain}$
		\EndIf
		\EndFor
		
		\State \textit{// Phase 2: Reinforcement and Restoration}
		\For{$id$ in $\mathcal{O}_k$}
		\If{$id \in \Omega.\text{Archived}$}
		\State $\Omega.\text{RestoreEntity}(id)$ \Comment{Restoration Event}
		\EndIf
		\State $v \leftarrow \Omega.\text{GetEntity}(id)$
		\State $v.confidence \leftarrow 1.0$ 
		\State $v.state \leftarrow \texttt{Active}$
		\EndFor
	\end{algorithmic}
\end{algorithm}

\textbf{Parameter Rationale:} We select $\lambda_{decay} = 0.95$. This implies a "half-life" of approx 13 scans. The archiving threshold $\tau_{archive}=0.1$ requires $\approx 45$ consecutive misses, providing a robust buffer against transient occlusions common in collaborative construction.

\subsection{Force-Torque Feedback Integration}
\label{app:force_feedback}

While VLM-DEWM focuses on cognitive-level planning (not low-level impedance control), the system interfaces with the robot controller's force-torque (F/T) signals to trigger state transitions in the $CS$ module. This section specifies the integration protocol.

\paragraph{a. Scope Clarification}
We do \textit{not} design force control laws. Instead, we treat F/T readings as \textbf{binary event triggers} that signal the success or failure of contact-rich actions to the cognitive layer. The underlying impedance/admittance control is handled by the robot's native controller (Franka Control Interface in our experiments).

\paragraph{b. Threshold-Based Event Detection}
For each contact-rich action type, we define conservative thresholds based on the Franka Emika Panda's specifications and empirical tuning:

\begin{table}[h]
	\centering
	\caption{Force-Torque Thresholds for State Transition Triggers}
	\label{tab:ft_thresholds}
	\small
	\begin{tabular}{llcc}
		\toprule
		\textbf{Action} & \textbf{Signal} & \textbf{Threshold} & \textbf{Triggered Event} \\
		\midrule
		Grasp & Gripper Force $F_g$ & $< 5$ N (after close) & Grasp Failure \\
		Grasp & Gripper Force $F_g$ & $> 40$ N & Over-squeeze (abort) \\
		Transport & End-effector $|F_{ext}|$ & $> 15$ N & Collision Detected \\
		Place & End-effector $F_z$ & $> 10$ N (sustained) & Contact Confirmed \\
		Insert & End-effector $|\tau_{ext}|$ & $> 2$ Nm & Jamming Detected \\
		\bottomrule
	\end{tabular}
\end{table}

\paragraph{c. Mapping to $CS$ State Transitions}
The F/T events are mapped to $CS$ updates as follows:

\begin{itemize}
	\item \textbf{Grasp Failure} ($F_g < 5$N after gripper close command): \\
	$CS$ remains at $\langle\texttt{Idle}, \texttt{NULL}\rangle$; ERT's $\mathcal{A}_{causal}$ is invalidated.
	
	\item \textbf{Collision Detected} ($|F_{ext}| > 15$N during transport): \\
	Motion halts; $CS$ transitions to $\langle\texttt{Interrupted}, \text{last\_target}\rangle$; triggers re-perception.
	
	\item \textbf{Contact Confirmed} ($F_z > 10$N sustained for 0.5s during place): \\
	$CS$ transitions from $\langle\texttt{Placing}, \text{obj}\rangle$ to $\langle\texttt{Idle}, \texttt{NULL}\rangle$; placement success logged.
\end{itemize}

\paragraph{d. Implementation Note}
These thresholds are \textbf{fixed} in our experiments, calibrated once for the Franka+Robotiq platform. For deployment on different hardware, re-calibration is required. Adaptive thresholding based on object properties (e.g., fragile objects use lower $F_g$ limits) is a direction for future work (Section~\ref{sec:limitations}).

\section{Implementation Details of Dynamic Context Engineering}
\label{app:context_engineering}

This appendix details the subgraph extraction algorithm and prompt synthesis logic for the context engineering pipeline described in Section~\ref{sec:context_engineering}.

\subsection{Task-Relevant Subgraph Extraction Algorithm}
\label{app:subgraph_extraction}

The ``Information Bottleneck'' principle in Section~\ref{sec:context_engineering} is operationalized through a \textbf{two-stage filtering} algorithm, not learned compression. We prioritize interpretability and determinism over optimality.

\subsubsection{Stage 1: Zone-Based Spatial Filtering}

Given the current subtask $ss_t$ and robot state $CS_t$, we first identify the \textbf{relevant zones}:
\begin{equation}
	\mathcal{Z}_{relevant} = \{Z_{current}\} \cup \{Z \in \mathcal{G}_{zone} \mid Z \text{ mentioned in } ss_t.\text{goal}\}
\end{equation}
This exploits the hierarchical structure of $\mathcal{G}$ to immediately prune objects in distant, task-irrelevant areas.

\textit{Example:} For $ss_t = $ ``Pick red\_block from Zone\_A'', we have $\mathcal{Z}_{relevant} = \{\text{Zone\_A}\}$, ignoring all objects in Zone\_B, Zone\_C, etc.

\subsubsection{Stage 2: Task-Driven Relevance Scoring}

Within $\mathcal{Z}_{relevant}$, we compute a relevance score for each vertex $v_i \in \mathcal{G}_{obj}$:
\begin{equation}
	\begin{split}
		r(v_i) &= w_1 \cdot \mathbb{I}[v_i \in ss_t.\text{args}] \\
		&\quad + w_2 \cdot \mathbb{I}[\exists e: (v_i, v_j) \land v_j \in ss_t.\text{args}] \\
		&\quad + w_3 \cdot \text{sim}(v_i.\text{cat}, ss_t.\text{keywords})
	\end{split}
\end{equation}
where:
\begin{itemize}
	\item $w_1 = 1.0$: Direct task arguments (e.g., the object to be picked)
	\item $w_2 = 0.5$: 1-hop neighbors of task arguments (e.g., the surface it's on)
	\item $w_3 = 0.2$: Semantic similarity to task keywords (e.g., ``container'' for a packing task)
	\item $\text{sim}(\cdot)$: Embedding cosine similarity (precomputed, not runtime VLM call)
\end{itemize}

\subsubsection{Subgraph Construction}

The extracted subgraph $G_{sub}$ is formed by selecting vertices that exceed a relevance threshold:
\begin{align}
	V_{sub} &= \{v_i \mid r(v_i) > \tau_{rel}\} \cup \{v_{robot}\} \\
	E_{sub} &= \{e_{ij} \in \mathcal{E} \mid v_i \in V_{sub} \land v_j \in V_{sub}\}
\end{align}
We set $\tau_{rel} = 0.3$ empirically. This value was chosen to effectively prune irrelevant background objects (typically $r < 0.2$) while retaining indirect dependencies (e.g., support surfaces with $r \approx 0.5$). This filtering typically yields $|V_{sub}| \approx 3$--$8$ vertices for standard manipulation tasks.

\subsubsection{Algorithm Pseudocode}

\begin{algorithm}[H]
	\caption{Task-Relevant Subgraph Extraction}
	\label{alg:subgraph}
	\begin{algorithmic}[1]
		\Require $\mathcal{G}$, $ss_t$, $CS_t$, threshold $\tau_{rel}$
		\Ensure $G_{sub} = (V_{sub}, E_{sub})$
		\State $\mathcal{Z}_{rel} \gets \textsc{GetRelevantZones}(ss_t, CS_t)$
		\State $V_{candidates} \gets \{v \in \mathcal{G}_{obj} \mid v.\text{zone} \in \mathcal{Z}_{rel}\}$
		\State $V_{sub} \gets \emptyset$
		\For{$v_i \in V_{candidates}$}
		\State $r_i \gets \textsc{ComputeRelevance}(v_i, ss_t)$ \Comment{Eq. above}
		\If{$r_i > \tau_{rel}$}
		\State $V_{sub} \gets V_{sub} \cup \{v_i\}$
		\EndIf
		\EndFor
		\State $E_{sub} \gets \{(v_i, v_j) \in \mathcal{E} \mid v_i, v_j \in V_{sub}\}$
		\State \Return $(V_{sub}, E_{sub})$
	\end{algorithmic}
\end{algorithm}

\subsubsection{Clarification on ``Information Bottleneck''}

We use the Information Bottleneck (IB) principle as a \textbf{conceptual motivation}, not a learned objective. Our rule-based extraction approximates the IB goal of minimizing $I(Z; \Omega)$ (irrelevant info) while maximizing $I(Z; Y)$ (task-relevant info) through:
\begin{itemize}
	\item Zone filtering $\approx$ spatial compression
	\item Relevance scoring $\approx$ task-conditional selection
\end{itemize}
A learned IB encoder (e.g., via variational methods) is left for future work.

\subsection{Initialization and Task Decomposition Context}
\label{app:task_decomposition}
\textbf{Objective:} To decompose a high-level user command into a structured Directed Acyclic Graph (DAG) of sub-tasks.

\paragraph{a. Data Extraction:} 
The system scans the initial environment $\Omega_0$ and compiles a list of available entities: $\mathcal{L}_{res} = \{ (id_i, cat_i, \mathbf{p}_i) \}$.

\paragraph{b. Prompt Template:}
\begin{tcolorbox}[colback=gray!5, colframe=black, title=Prompt: Task Decomposition]
	\textbf{Role:} Expert Robotic Planner. \\
	\textbf{User Command:} "\{T\}" \\
	\textbf{Available Resources:} \\
	- \{resource\_list\} \\
	\textbf{Objective:} Decompose the command into a sequence of sub-tasks. Output a JSON object representing a DAG, where each node contains: "id", "function" (e.g., pick, place), "args", and "dependencies". \\
	\textbf{Example Output:} \\
	\{ "nodes": [ \{ "id": 1, "function": "pick", "args": ["blue\_block\_1"], "dependencies": [] \} ] \}
\end{tcolorbox}

\subsection{Coarse Motion Planning Context}
\label{app:coarse_planning}
\textbf{Objective:} To generate collision-free waypoints for macro-navigation.

\paragraph{a. Data Extraction:}
We cluster the background map $\mathcal{S}_{bg}$ (e.g., using DBSCAN) to extract obstacle bounding boxes $\mathcal{O}_{list}$.

\paragraph{b. Prompt Template:}
\begin{tcolorbox}[colback=gray!5, colframe=black, title=Prompt: Coarse Path Planning]
	\textbf{Current Goal:} "\{SS\_t.description\}" \\
	\textbf{Robot State:} \{CS\_t\} \\
	\textbf{Constraints:} \\
	- Obstacles (BBoxes): \{obstacle\_list\} \\
	- Semantic Constraints: \{relevant\_relations\_from\_G\} \\
	\textbf{Task:} Generate a list of SE(3) waypoints for the end-effector to reach the target area while avoiding obstacles.
\end{tcolorbox}

\subsection{Fine-Grained Manipulation Context}
\label{app:fine_manipulation}
\textbf{Objective:} To generate precise delta-pose adjustments for contact-rich interaction.

\paragraph{a. Data Extraction:}
We query high-fidelity data sources: Shape Prior $\mathcal{L}$, Force-Torque sensors $F_t$, and precise object pose $\mathbf{T}_{obj}$.

\paragraph{b. Prompt Template:}
\begin{tcolorbox}[colback=gray!5, colframe=black, title=Prompt: Fine Manipulation]
	\textbf{Target Object:} \{target.id\} (\{target.category\}) \\
	- Pose (SE3): \{target.pose\} \\
	- Envelope ($\mu, \Sigma$): \{target.envelope\} \\
	- Shape Prior: \{shape\_desc\} \\
	- Force Readings: \{force\_torque\_data\} \\
	\textbf{Task:} Based on the geometry and force feedback, compute the next optimal delta-pose ($\Delta \mathbf{T}$) for the end-effector. Return a JSON with "action\_type" and "value".
\end{tcolorbox}

\subsection{Failure Diagnosis and Recovery Context}
\label{app:failure_diagnosis}
\textbf{Objective:} To analyze execution failures and propose a recovery plan.

\paragraph{a. Data Extraction:}
We compile the Action Sequence History $\mathcal{H}_{ash}$ and the post-failure world state $\Omega_{fail}$.

\paragraph{b. Prompt Template:}
\begin{tcolorbox}[colback=gray!5, colframe=black, title=Prompt: Failure Diagnosis]
	\textbf{CRITICAL FAILURE REPORT} \\
	\textbf{1. Failed Action:} "\{failed\_SS\_t.description\}" \\
	\textbf{2. Context:} \\
	- State Before: \{CS\_t\_before\} \\
	- Action Log: \{recent\_action\_log\} \\
	\textbf{3. Post-Mortem:} \\
	- Error Code: \{failure\_reason\} \\
	- Current World State: \{summary\_of\_S\_and\_G\} \\
	\textbf{Task:} \\
	1. Diagnose the root cause of the failure. \\
	2. Propose a new TTP (DAG JSON) to recover and achieve the original goal.
\end{tcolorbox}

\subsection{Formal Definition of Semantic Discrepancy Operator ($\ominus$)}
\label{app:discrepancy_operator}

The semantic discrepancy $\Delta = \Omega_{t+1} \ominus \mathcal{A}_{causal}$ in Section~\ref{sec:deep_diagnosis} is computed via a \textbf{rule-based state comparison}, not probabilistic VLM inference. This ensures deterministic and auditable diagnosis.

\paragraph{a. Definition}
Given the expected post-state $\mathcal{A}_{causal} = \langle CS_{exp}, \mathcal{R}_{exp} \rangle$ (where $CS_{exp}$ is the expected constraint state and $\mathcal{R}_{exp}$ is a set of expected relation changes) and the actual observed state $\Omega_{t+1}$, we define:
\begin{equation}
	\Delta = \Omega_{t+1} \ominus \mathcal{A}_{causal} := \langle \Delta_{CS}, \Delta_{\mathcal{G}}, \Delta_{\mathcal{S}} \rangle
\end{equation}

\paragraph{b. Component-wise Computation:}
\textbf{b.1. Constraint State Discrepancy ($\Delta_{CS}$):}
A direct symbolic comparison between expected and actual micro-topological states:
\begin{equation}
	\Delta_{CS} = 
	\begin{cases}
		\texttt{NULL} & \text{if } CS_{t+1} = CS_{exp} \\
		(CS_{exp}, CS_{t+1}) & \text{otherwise}
	\end{cases}
\end{equation}
\textit{Example:} If $CS_{exp} = \langle\texttt{Holding}, \text{obj\_1}\rangle$ but $CS_{t+1} = \langle\texttt{Idle}, \texttt{NULL}\rangle$, then $\Delta_{CS} = (\texttt{Holding} \to \texttt{Idle})$, indicating a grasp failure.

\textbf{b.2. Semantic Graph Discrepancy ($\Delta_{\mathcal{G}}$):}
We perform a structural graph difference on the object-level semantic graph:
\begin{align}
	\Delta_{\mathcal{G}}^{+} &= \mathcal{E}_{t+1} \setminus \mathcal{E}_{exp} \quad \text{(unexpected new edges)} \\
	\Delta_{\mathcal{G}}^{-} &= \mathcal{E}_{exp} \setminus \mathcal{E}_{t+1} \quad \text{(missing expected edges)}
\end{align}
where $\mathcal{E}$ denotes the edge set of $\mathcal{G}_{obj}$. This is implemented as a deterministic set operation on relation tuples.

\textit{Example:} If the action was \texttt{Place(A, B)} with $\mathcal{R}_{exp} = \{+\texttt{On}(A,B)\}$, but post-observation shows $\texttt{On}(A, \texttt{floor})$, then:
\begin{itemize}
	\item $\Delta_{\mathcal{G}}^{-} = \{\texttt{On}(A,B)\}$ (expected edge missing)
	\item $\Delta_{\mathcal{G}}^{+} = \{\texttt{On}(A,\texttt{floor})\}$ (unexpected edge present)
\end{itemize}

\textbf{b.3. Geometric Discrepancy ($\Delta_{\mathcal{S}}$):}
For objects involved in the action, we compute the pose deviation:
\begin{equation}
	\Delta_{\mathcal{S}}(o_i) = \|\boldsymbol{\mu}_{t+1}(o_i) - \boldsymbol{\mu}_{exp}(o_i)\|_2
\end{equation}
A significant deviation ($> \tau_{geo\_drift}$, default 0.05m) flags potential unmodeled disturbances.

\paragraph{c. Discrepancy Report Generation:}
The structured $\Delta$ is serialized into natural language for VLM-based recovery planning:

\begin{tcolorbox}[colback=gray!5, colframe=black, title=Discrepancy Report Template]
	\textbf{DISCREPANCY DETECTED} \\
	- CS Mismatch: Expected \{$CS_{exp}$\}, Actual \{$CS_{t+1}$\} \\
	- Missing Relations: \{$\Delta_{\mathcal{G}}^{-}$\} \\
	- Unexpected Relations: \{$\Delta_{\mathcal{G}}^{+}$\} \\
	- Geometric Drift: \{object: drift\_distance\} \\
	\textbf{Likely Cause:} [Inferred from $\Delta_{CS}$ pattern, see Table~\ref{tab:diagnosis_rules}]
\end{tcolorbox}

\paragraph{d. Diagnosis Rule Table:}
The $\Delta_{CS}$ pattern deterministically maps to failure categories:

\begin{table}[h]
	\centering
	\caption{Rule-Based Failure Diagnosis from $\Delta_{CS}$}
	\label{tab:diagnosis_rules}
	\small
	\begin{tabular}{lll}
		\toprule
		\textbf{$CS_{exp}$} & \textbf{$CS_{t+1}$} & \textbf{Diagnosed Failure} \\
		\midrule
		Holding(X) & Idle & Grasp Slip / Gripper Failure \\
		Holding(X) & Holding(Y) & Object Misidentification \\
		Approaching(X) & Idle & Motion Interrupted / Collision \\
		Placing(X, Y) & Holding(X) & Placement Rejected / Unstable \\
		\bottomrule
	\end{tabular}
\end{table}

This rule-based approach ensures that the $\ominus$ operator produces \textbf{deterministic, explainable} discrepancy reports, while the subsequent VLM call (\ref{app:failure_diagnosis}) is responsible only for \textbf{recovery strategy synthesis}—not diagnosis itself.

\section{Implementation Details of the ERT Validation Protocol}
\label{app:ert_validation}

This appendix details the schema and verification logic for the External Reasoning Trace (ERT) described in Section~\ref{sec:ert}.

\subsection{ERT Schema Definition}
The VLM output must strictly adhere to the schema defined in Table~\ref{tab:ert_schema}.

\begin{table}[h]
	\centering
	\caption{Schema Definition for External Reasoning Trace (ERT)}
	\label{tab:ert_schema}
	\small
	\begin{tabular}{l l p{7cm}}
		\toprule
		\textbf{Field} & \textbf{Type} & \textbf{Description \& Example} \\
		\midrule
		\texttt{action\_proposal} & String & The proposed skill call. \newline \textit{Ex: "Grasp(blue\_block\_1)"} \\
		\texttt{world\_belief} & Object & Structured assertion about state. \newline \textit{Ex: \{"pred": "On", "args": ["A", "B"]\}} \\
		\texttt{causal\_assump} & String & Expected outcome. \newline \textit{Ex: "CS becomes Holding(A)"} \\
		\texttt{confidence} & Float & Self-evaluated score [0.0, 1.0]. \\
		\texttt{fallback} & String & Contingency plan. \\
		\bottomrule
	\end{tabular}
\end{table}

\subsection{Multi-Layered Validation Logic}
The \texttt{GetValidERT()} function enforces a three-stage check:

\paragraph{a. Syntactic Validation:}
We use a JSON schema validator to ensure the output contains all required fields (Table~\ref{tab:ert_schema}) and that data types are correct (e.g., $0.0 \le \texttt{confidence} \le 1.0$).

\paragraph{b. Semantic Validation:}
We extract all entity IDs from \texttt{action\_proposal} and \texttt{world\_belief}. The validator queries the DEWM to ensure existence:
\begin{equation}
	\forall id \in \mathcal{E}_{ert}, \quad (id \in \mathcal{G}_{obj}) \lor (id \in \mathcal{G}_{zone})
\end{equation}
This prevents hallucinations of non-existent objects.

\paragraph{c. Physical Consistency Validation:}
We verify the \texttt{world\_belief} against the geometric ground truth $\mathcal{S}$.
\begin{itemize}
	\item \textbf{Verify On(A, B):}
	\begin{enumerate}
		\item \textit{Support:} Check if $A$'s centroid $\boldsymbol{\mu}_A$ projected onto the XY plane lies within $B$'s XY-envelope.
		\item \textit{Contact:} Check if the vertical gap is minimal: $|Z_{min}(A) - Z_{max}(B)| < \epsilon_{contact}$.
	\end{enumerate}
	\item \textbf{Verify Inside(A, B):}
	Check if $A$'s bounding box is fully contained within $B$'s boundaries:
	\begin{equation}
		(\mathbf{min}_A \ge \mathbf{min}_B) \land (\mathbf{max}_A \le \mathbf{max}_B)
	\end{equation}
	\item \textbf{Verify Near(A, B):}
	Check Euclidean distance: $\| \boldsymbol{\mu}_A - \boldsymbol{\mu}_B \| < d_{near}$.
\end{itemize}

\subsection{System Briefing (Meta-Instruction)}
To ensure the VLM utilizes the DEWM correctly, the following system prompt is injected at the start of every session:

\begin{tcolorbox}[colback=gray!5, colframe=black, title=System Briefing]
	\textbf{SYSTEM INSTRUCTION:} \\
	You are interacting with the \textbf{VLM-DEWM} system. This system maintains a persistent World Model with the following queryable modules: \\
	1. \textbf{Semantic Graph (G):} Tracks entity IDs, states (e.g., "OnTable"), and relations. \\
	2. \textbf{Spatial Network (S):} Maintains precise poses and geometry (Point Clouds/Voxels). \\
	3. \textbf{Task Memory (M):} Tracks current sub-task and dependency graph (TTP). \\
	4. \textbf{Robot State (CS):} Tracks interaction status (e.g., "Holding X"). \\
	\textbf{RULE:} Do not hallucinate state. Do not re-infer facts from raw images if they are already provided in the context. Treat the provided DEWM data as the Single Source of Truth.
\end{tcolorbox}

\section{Computational Complexity Analysis}
\label{app:complexity}

This appendix analyzes the computational complexity of the VLM-DEWM framework. The overall cost comprises three main components: (1) perception and world model update, (2) dynamic context engineering, and (3) VLM inference and validation. We assume the DEWM maintains $N_{obj}$ object instances, and each perception snapshot $Env_t$ observes $N_{obs}$ objects.

\subsection{Per-Cycle Update Cost}

This is the foundational cost incurred in each cognitive cycle.

\paragraph{a. Symbolic Perception Interface}
The VLM-driven semantic focusing requires one VLM call. Geometric processing (point cloud segmentation, projection, voxelization) scales as $O(N_{points})$ with the input point cloud size. Due to task-driven focusing, the actual processed point cloud is significantly smaller than the full scene.

\paragraph{b. Data Association}
This is one of the main bottlenecks in the update process.
\begin{itemize}
	\item \textbf{Geometric Candidate Screening:} A naive implementation comparing $N_{obs}$ observations against $N_{obj}$ memory entries has complexity $O(N_{obs} \cdot N_{obj})$. In our implementation, by leveraging the Hierarchical Semantic Graph (HSG) for spatial indexing (first matching to Zone) and octree-based nearest neighbor search on the Spatial Network $\mathcal{S}$, the average complexity is reduced to $O(N_{obs} \log N_{obj})$.
	\item \textbf{VLM Semantic Adjudication:} For each observation, we select from $k$ geometric candidates (typically $k \ll N_{obj}$), requiring $N_{obs} \cdot k$ VLM scoring calls (optimizable via batching), or a single VLM call containing all candidates.
\end{itemize}

\paragraph{c. Geometric Layer Update (S-Update)}
For each successfully associated object (at most $N_{obs}$), the Gaussian envelope fusion (Equations~\ref{eq:sigma_fusion}--\ref{eq:mu_fusion}) is $O(1)$. When high-precision pose estimation is required, the scale-aware ICP algorithm has complexity $O(I \cdot N_p \log N_p)$, where $I$ is the number of iterations and $N_p$ is the number of points in the object's local point cloud.

\paragraph{d. Semantic Layer Update (G-Update)}
VLM relation hypothesis generation requires one VLM call. The geometric verification step checks object-pair relations; in the worst case, this could involve $O(N_{obj}^2)$ pairs, but in practice, the VLM typically proposes only a constant number of the most relevant relation hypotheses. Each hypothesis verification (e.g., Mahalanobis distance computation) is $O(1)$.

\subsection{Context Engineering Cost}

The goal of this process is to extract task-relevant information from the DEWM.

\paragraph{a. Subgraph Extraction}
Thanks to the hierarchical structure of HSG, we first perform $O(1)$ indexing to identify the relevant Zone nodes. This restricts the search space from the global graph size $|V_{global}|$ to the local zone size $|V_{zone}|$ (where $|V_{zone}| \ll |V_{global}|$). Consequently, the relevance scoring involves iterating only over $|V_{zone}|$, while the text serialization complexity scales with the much smaller final subgraph size $O(|V_{sub}| + |E_{sub}|)$. Since $|V_{zone}|$ is physically bounded (i.e., independent of the total facility scale), the overall context engineering overhead remains $O(1)$ relative to the global DEWM scale.

\subsection{VLM-Guided Planning Cost}

\paragraph{a. VLM Inference}
VLM inference is the highest-latency component of the entire framework. Its complexity depends on the model architecture (typically Transformer) and input/output lengths. We can approximate it as $f(L_{prompt}, L_{gen})$, where $L_{prompt}$ is the token length of the context and prompt, and $L_{gen}$ is the token length of the generated ERT. A core objective of dynamic context engineering is to minimize $L_{prompt}$.

\paragraph{b. ERT Validation}
\begin{itemize}
	\item \textbf{Syntactic Validation:} JSON schema validation scales linearly with ERT length, $O(L_{gen})$.
	\item \textbf{Semantic Validation:} Checking whether entities in the ERT exist in $\mathcal{G}$. If $\mathcal{G}$'s vertices are stored using a hash table, each lookup has average complexity $O(1)$.
	\item \textbf{Physical Consistency Validation:} Checking $world\_belief$ against $\mathcal{S}$ typically involves $O(1)$ geometric computations.
\end{itemize}

\subsection{Summary}

The main computational bottlenecks are \textbf{VLM inference latency} and \textbf{data association overhead in densely cluttered scenes}. Our architectural design, particularly the \textbf{Hierarchical Semantic Graph (HSG)} and \textbf{Dynamic Context Engineering}, aims to significantly prune the search space through spatial and task context, thereby effectively controlling the growth of data association cost and VLM context length. This enables the system to maintain high operational efficiency even when facing long-term, complex tasks.

Table~\ref{tab:complexity_summary} provides a summary of the complexity analysis.

\begin{table}[H]
	\centering
	\caption{Summary of Computational Complexity}
	\label{tab:complexity_summary}
	\small
	\begin{tabular}{ll}
		\toprule
		\textbf{Component} & \textbf{Complexity} \\
		\midrule
		Data Association (naive) & $O(N_{obs} \cdot N_{obj})$ \\
		Data Association (with HSG) & $O(N_{obs} \log N_{obj})$ \\
		Gaussian Envelope Fusion & $O(1)$ per object \\
		Scale-aware ICP & $O(I \cdot N_p \log N_p)$ \\
		Subgraph Extraction & $O(|V_{zone}|)$ \\
		ERT Semantic Validation & $O(1)$ per entity \\
		VLM Inference & $f(L_{prompt}, L_{gen})$ \\
		\bottomrule
	\end{tabular}
	\footnotesize{\textit{Note: $|V_{zone}|$ denotes the number of entities in the relevant topological zone, which is physically bounded.}}
\end{table}

\bibliography{references.bib}

\end{document}